%% file: neurips_2021.tex
\documentclass{article}

\usepackage[final]{neurips_2021}

\usepackage[utf8]{inputenc} 
\usepackage[T1]{fontenc}    
\usepackage{hyperref}       
\usepackage{url}            
\usepackage{booktabs}       
\usepackage{amsfonts}       
\usepackage{nicefrac}       
\usepackage{microtype}      
\usepackage{xcolor}         
\usepackage{graphicx} 
\usepackage{caption}
\usepackage{wrapfig}
\usepackage{algorithmic}
\usepackage{tikz}
\usetikzlibrary{arrows,positioning,automata,shadows,fit,shapes}
\usepackage{float}
\usepackage{subfig}

\usepackage{bookmark}
\usepackage{natbib}
\usepackage{amsmath}
\usepackage{amssymb}
\usepackage{dsfont}
\usepackage[ruled,vlined,linesnumbered]{algorithm2e}
\usepackage{colortbl}
\usepackage{chemfig}
\usepackage{bm}
\usepackage{placeins}
\usepackage{eqnarray}
\usepackage{multirow}

\include{latex-math/basic-math}
\include{latex-math/basic-ml}
\include{latex-math/ml-mbo}

\title{Explaining Hyperparameter Optimization\\ via Partial Dependence Plots}

\author{Julia Moosbauer$\thanks{These authors contributed equally to this work.}$~, Julia Herbinger$^{\ast}$, Giuseppe Casalicchio, Marius Lindauer, Bernd Bischl \\
Department of Statistics,
Ludwig-Maximilians-University Munich, Munich, Germany\\
Institute of Information Processing, Leibniz University Hannover, Hannover, Germany \\
\texttt{\{julia.moosbauer, julia.herbinger, giuseppe.casalicchio, }\\ \texttt{bernd.bischl\}@stat.uni-muenchen.de}\\
\texttt{lindauer@tnt.uni-hannover.de}
}

\begin{document}

\maketitle

\begin{abstract}
Automated hyperparameter optimization (HPO) can support practitioners to obtain peak performance in machine learning models.
However, there is often a lack of valuable insights into the effects of different hyperparameters on the final model performance.
This lack of explainability makes it difficult to trust and understand the automated HPO process and its results.
We suggest using interpretable machine learning (IML) to gain insights from the experimental data obtained during HPO with Bayesian optimization (BO).
BO tends to focus on promising regions with potential high-performance configurations and thus induces a sampling bias.
Hence, many IML techniques, such as the \emph{partial dependence plot} (PDP), carry the risk of generating biased interpretations.
By leveraging the posterior uncertainty of the BO surrogate model, we introduce a variant of the PDP with estimated confidence bands.
We propose to partition the hyperparameter space to obtain more confident and reliable PDPs in relevant sub-regions.
In an experimental study, we provide quantitative evidence for the increased quality of the PDPs within sub-regions.
\end{abstract}

\section{Introduction}
\label{sec:introduction}

Most machine learning (ML) algorithms are highly configurable. 
Their hyperparameters must be chosen carefully, as their choice often impacts the model performance. 
Even for experts, it can be challenging to find well-performing hyperparameter configurations. 
Automated machine learning (AutoML) systems and methods for automated HPO have been shown to yield considerable efficiency compared to manual tuning by human experts \citep{snoek2012practical}.
However, these approaches mainly return a well-performing configuration
and leave users without insights into decisions of the optimization process.
Questions about the importance of hyperparameters or their effects on the resulting performance often remain unanswered.
Not all data scientists trust the outcome of an AutoML system due to the lack of transparency \citep{drozdal2020trust}.
Consequently, they might not deploy an AutoML model, despite all performance gains. 
Providing insights into the search process may help increase trust and facilitate interactive and exploratory processes: A data scientist could monitor the AutoML process and make changes to it (e.g., restricting or expanding the search space) already \emph{during} optimization to anticipate unintended results.

Transparency, trust, and understanding of the inner workings of an AutoML system can be increased by interpreting the internal surrogate model of an AutoML approach. 
For example, BO trains a surrogate model to approximate the relationship between hyperparameter configurations and model performance. It is used to guide the optimization process towards the most promising regions of the hyperparameter space.
Hence, surrogate models implicitly contain information 
about the influence of hyperparameters. If the interpretation of the surrogate matches with a data scientist's expectation, confidence in the correct functioning of the system may be increased. If these do not match, it provides an opportunity to look either for bugs in the code or for new theoretical insights.

We propose to analyze surrogate models with methods from IML to provide insights into the results of HPO. 
In the context of BO, typical choices for surrogate models are flexible, probabilistic black-box models, such as Gaussian processes (GP) or random forests.
Interpreting the effect of single hyperparameters on the performance of the model to be tuned is analogous to interpreting the feature effect of the black-box surrogate model.
We focus on the PDP \citep{friedman2001greedy}, which is a widely-used method\footnote{There exist various implementations \citep{greenwell:2017, scikit-learn}), extensions \citep{greenwell2018simple, goldstein2014peeking} and applications \citep{friedman2003multiple, cutler2007random}.}  to visualize the average marginal effect of single features on a black-box model's prediction.
When applied to surrogate models, they provide information on how a specific hyperparameter influences the estimated model performance. 
However, applying PDPs out of the box to surrogate models might lead to misleading conclusions.
Efficient optimizers such as BO tend to focus on exploiting promising regions of the hyperparameter space while leaving other regions less explored. Therefore, a sampling bias in input space is introduced, which in turn can lead to a poor fit and biased interpretations in underexplored regions of the space.

\textbf{Contributions:}
We study the problem of sampling bias in experimental data produced by AutoML systems and the resulting bias of the surrogate model and assess its implications on PDPs.
We then derive an uncertainty measure for PDPs of probabilistic surrogate models.
In addition, we propose a method that splits the hyperparameter space into interpretable sub-regions of varying uncertainty to obtain sub-regions with more reliable and confident PDP estimates.
In the context of BO, we provide evidence for the usefulness of our proposed methods on a synthetic function and in an experimental study in which we optimize the architecture and hyperparameters of a deep neural network.
Our Supplementary Material provides (A) more background related to uncertainty estimates, (B) notes on how our methods are applied to hierarchical hyperparameter spaces, (C) details on the experimental setup and more detailed results, (D) a link to the source code.

\textbf{Reproducibility and Open Science}: The implementation of the proposed methods as well as reproducible scripts for the experimental analysis are provided in a public git-repository\footnote{\url{https://github.com/slds-lmu/paper_2021_xautoml}}.

\section{Background and Related Work}
\label{sec:background}

Recent research has begun to question whether the evaluation of an AutoML system should be purely based on the generated models' predictive performance without considering interpretability \citep{hutter14, pfisterer:2019, Freitas:2019, xanthopoulos:2020}.
Interpreting AutoML systems can be categorized as (1)~interpreting the resulting ML model on the underlying dataset, or (2) interpreting the HPO process itself.
In this paper, we focus on the latter. 

Let $c: \Lambda \to \R$ be a \emph{black-box} cost function, mapping a hyperparameter configuration \mbox{$\lambdab = \left(\lambda_1, ..., \lambda_d\right)$} to the model error\footnote{Typically, the model error is estimated via cross-validation or hold-out testing.} obtained by a learning algorithm with configuration $\lambdab$. The hyperparameter space may be mixed, containing categorical and continuous hyperparameters. 
The goal of HPO is to find $\lambdab^\ast \in \argmin\nolimits_{\lambdab \in \Lambda} c(\lambdab).$
Throughout the paper, we assume that a surrogate model $\hat c: \Lambda \to \R$ is given as an approximation to $c$. 
If the surrogate is assumed to be a GP, $\hat c(\lambdab)$ is a random variable following a Gaussian posterior distribution. In particular, for any finite indexed family of hyperparameter configurations $\left(\lambdab^{(1)}, ..., \lambdab^{(k)} \right) \in \Lambda^k$, the vector of estimated performance values is Gaussian with a posterior mean $\bm{\hat m} = \left(\hat m\left(\lambdab^{(i)}\right)\right)_{i = 1, ..., k}$ and covariance $\bm{\hat K} = \left(\hat k\left(\lambdab^{(i)}, \lambdab^{(j)}\right)\right)_{i, j = 1, ..., k}$. 

\textbf{Hyperparameter Importance.} 
Understanding which hyperparameters influence model performance can provide valuable insights into the tuning strategy \citep{probst2018tunability}.
To quantify relevance of hyperparameters, models that inherently quantify feature relevance -- e.g., GPs with ARD kernel \citep{neil1996bayesian} -- can be used as surrogate models.  
\citet{hutter14} quantified the importance of hyperparameters based on a random forest fitted on data generated by BO, for which the importance of both the main and the interaction effects of hyperparameters was calculated by a functional ANOVA approach. Similarly, \citet{sharma:2019} quantified the hyperparameter importance of residual neural networks.
These works highlight how useful it is to quantify the importance of hyperparameters. 
However, importance scores do not show \emph{how} a specific hyperparameter affects the model performance according to the surrogate model. Therefore, we propose to visualize the assumed marginal effect of a hyperparameter. 
A model-agnostic interpretation method that can be used for this purpose is the PDP.


\textbf{PDPs for Hyperparameters.}
Let $S\subset \{1, 2, ..., d\}$ denote an index set of features, and let $C = \{1, 2, ..., d\} \setminus S$ be its complement. 
The partial dependence (PD) function \citep{friedman2001greedy} of $c: \Lambda \to \R$ for hyperparameter(s) $S$ is defined as\footnote{To keep notation simple, we denote $c(\lambdab)$ as a function of two arguments $(\lambdab_S, \lambdab_C)$ to differentiate components in the index set $S$ from those in the complement. The integral shall be understood as a multiple integral of $c$ where $\lambdab_j$, $j \in C$, are integrated out. }
\begin{eqnarray}
       c_{S}(\lambdab_S) := \E_{\lambdab_C}\left[c(\lambdab)\right]=\int_{\Lambda_C} c(\lambdab_S, \lambdab_C)~\textrm{d}\mathbb{P}(\lambdab_C).
       \label{eq:pdp}
\end{eqnarray}
When analyzing the PDP of hyperparameters, we are usually interested in how their values $\lambdab_S$ impact model performance uniformly across the hyperparameter space. In line with prior work~\citep{hutter14}, we therefore assume $\P$ to be the uniform distribution over $\Lambda_C$.
Computing $c_S(\lambdab_S)$ exactly is usually not possible because $c$ is unknown and expensive to evaluate in the context of HPO. Thus, the posterior mean $\hat{m}$ of the probabilistic surrogate model $\hat{c}(\lambdab)$ is commonly used as a proxy for $c$. Furthermore, the integral may not be analytically tractable for arbitrary surrogate models $\hat c$. Hence, the integral is approximated by Monte Carlo integration, i.e.,
\begin{eqnarray}
    \hat c_S\left(\lambdab_S\right) &=& \frac{1}{n} \sum\nolimits_{i = 1}^n \hat m\left(\lambdab_S, \lambdab_C^{(i)}\right)
    \label{eq:estimate_pdp}
\end{eqnarray}
for a sample $\left(\lambdab_C^{(i)}\right)_{i = 1, ..., n} \sim \P(\lambdab_C)$. $\hat m \left(\lambdab_S, \lambdab_C^{(i)}\right)$ represents the marginal effect of $\lambdab_S$ for one specific instance $i$. 
Individual conditional expectation (ICE) curves \citep{goldstein2014peeking} visualize the marginal effect of the $i$-th observation by plotting the value of $\hat m \left(\lambdab_S, \lambdab_C^{(i)}\right)$ against $\lambdab_S$ for a set of grid points\footnote{Grid points are typically chosen as an equidistant grid or sampled from $\P(\lambdab_S)$. The granularity $G$ is chosen by the user. For categorical features, the granularity typically corresponds to the number of categories.}  $\lambdab_S^{(g)}\in \Lambda_S$, $g \in \{1, ..., G\}$.
Analogously, the PDP visualizes $\hat c_{S}(\lambdab_S)$ against the grid points.
Following from Eq.~\ref{eq:estimate_pdp}, the PDP 
visualizes the average over all ICE curves.
In HPO, the marginal predicted performance is a related concept. Instead of approximating the integral via Monte Carlo, the integral over $\hat c$ is computed exactly. 
\citet{hutter14} propose an efficient approach to compute this integral for random forest surrogate models.

\textbf{Uncertainty Quantification in PDPs.} 
Quantifying the uncertainty of PDPs provides additional information about the reliability of the mean estimator. 
\citet{hutter14} quantified the model uncertainty specifically for random forests as surrogates in BO by calculating the standard deviation of the marginal predictions of the individual trees.
However, this procedure is not applicable to general probabilistic surrogate models, such as the commonly used GP.
There are approaches that quantify the uncertainty for ML models that do not provide uncertainty estimates out-of-the-box. 
\citet{cafri:2016} suggested a bootstrap approach for tree ensembles to quantify the uncertainties of effects based on PDPs.
Another approach to quantify the uncertainty of PDPs is to leverage the ICE curves. For example, \citet{greenwell:2017} implemented a method that marginalizes over the mean $\pm$ standard deviation of the ICE curves for each grid point. 
However, this approach quantifies the underlying uncertainty of the data at hand rather than the model uncertainty, as explained in Appendix \ref*{sec:app_choice_uncertainty}. A model-agnostic estimate based on uncertainty estimates for probabilistic models is missing so far. 

\textbf{Subgroup PDPs.}
Recently, a new research direction concentrates on finding more reliable PDP estimates within subgroups of observations. \citet{molnar2020modelagnostic} focused on problems in PDP estimation with correlated features. To that end, they apply transformation trees to find homogeneous subgroups and then visualize a PDP for each subgroup. \citet{groemping:2020} looked at the same problem and also uses subgroup PDPs, where ICE curves are grouped regarding a correlated feature. \citet{britton:2019} applied a clustering approach to group ICE curves to find interactions between features. 
However, none of these approaches aim at finding subgroups where reliable PDP estimates have low uncertainty. 
Additionally, to the best of our knowledge, nothing similar exists for analyzing experimental data created by HPO. 

\section{Biased Sampling in HPO}
\label{sec:bias}
\begin{figure}[t]
    \centering
    \begin{minipage}[t]{.48\textwidth}
        \includegraphics[width=\textwidth]{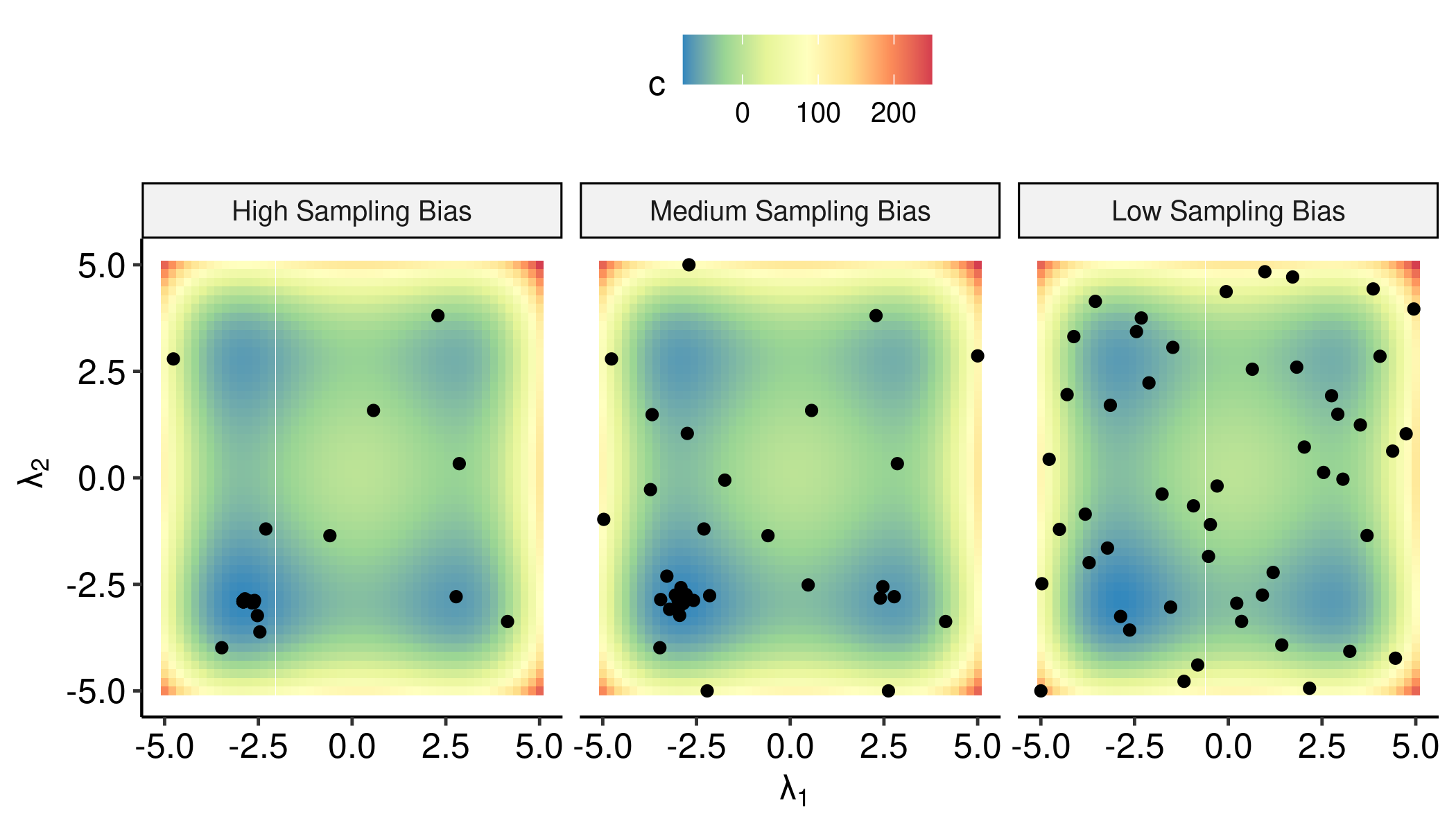}
        \caption{Illustration of the sampling bias when optimizing the $2D$ Styblinski Tang function with BO and the Lower Confidence Bound (LCB) acquisition function $a(\lambdab) = \hat m(\lambdab) + \tau \cdot \hat s (\lambdab)$ for $\tau = 0.1$ (left) and $\tau = 2$ (middle) vs. data sampled uniformly at random (right). }
        \label{fig:sampling_bias}
    \end{minipage}%
    \hfill
    \begin{minipage}[t]{0.48\textwidth}
        \centering
        \includegraphics[width=\textwidth]{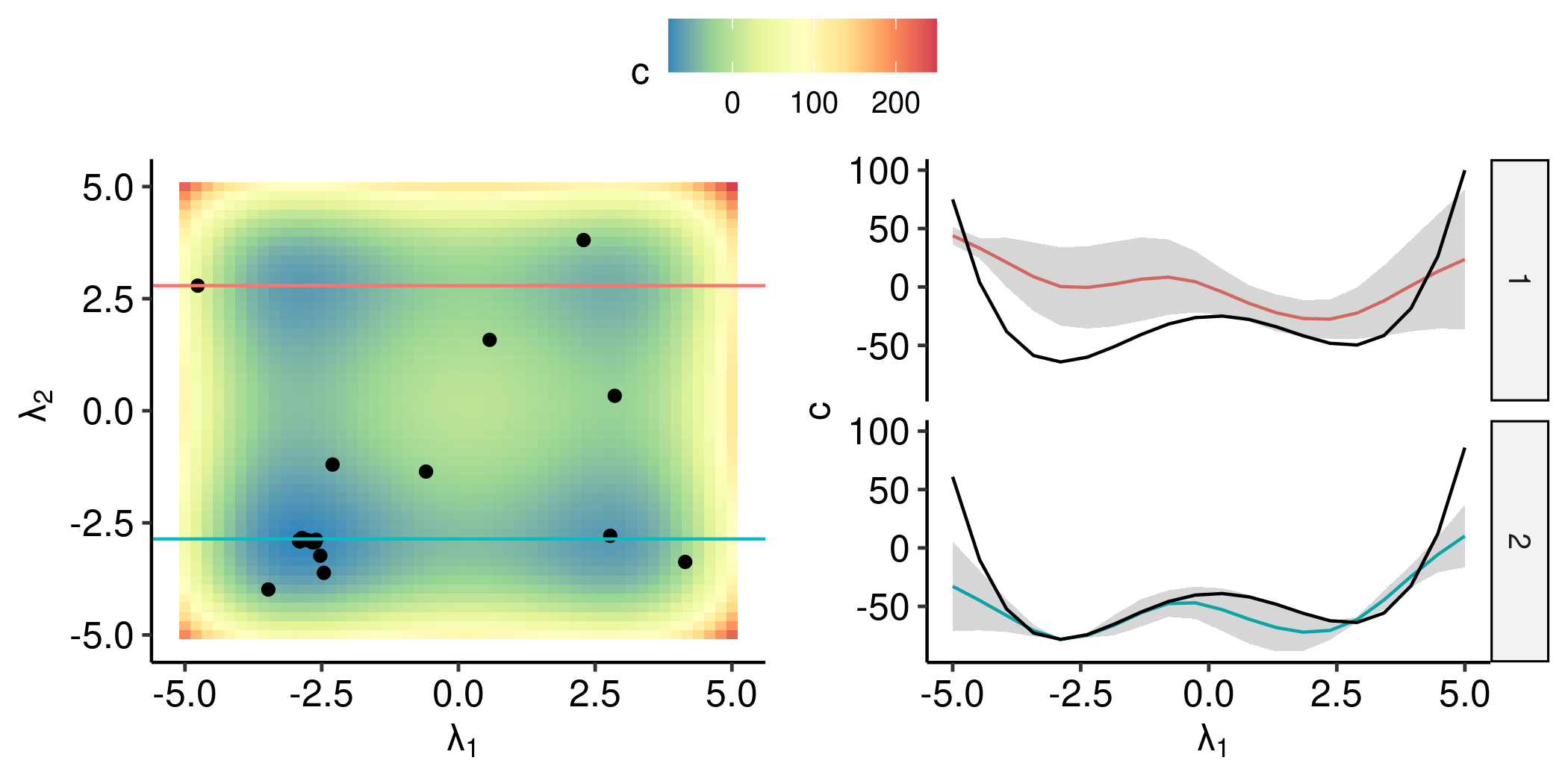}
        \caption{The two horizontal cuts (left) yield two ICE curves (right) showing the mean prediction and uncertainty band against $\lambda_1$ for $\hat c$ with $\tau = 0.1$ on the $2D$ Styblinski-Tang function.
        The upper ICE curve deviates more from the true effect (black) and shows a higher uncertainty. 
        }
        \label{fig:uncertainty_ice_curves}
    \end{minipage}
\end{figure}

Visualizing the marginal effect of hyperparameters of surrogate models via
PDPs can be misleading.
We show that this problem is due to the sequential nature of BO, which generates dependent instances (i.e., hyperparameter configurations) and thereby introduces a sampling and a resulting model bias.
To save computational resources in contrast to grid search or random search, efficient optimizers like BO tend to exploit promising regions of the hyperparameter space while other regions are less explored (see Figure \ref{fig:sampling_bias}). 
Consequently, predictions of surrogate models are usually more accurate with less uncertainty in well-explored regions and less accurate with high uncertainty in under-explored regions.
This model bias also affects the PD estimate (see Figure \ref{fig:uncertainty_ice_curves}). 
ICE curves may be biased and less confident if they are computed in poorly-learned regions where the model has not seen much data before.
Under the assumption of uniformly distributed hyperparameters, poorly-learned regions are incorporated in the PD estimate with the same weight as well-learned regions. 
ICE curves belonging to regions with high uncertainty may obfuscate well-learned effects of ICE curves belonging to other regions when they are aggregated to a PDP.
Hence, the model bias may also lead to a less reliable PD estimate. 
PDPs visualizing only the mean estimator of Eq. \eqref{eq:estimate_pdp} do not provide insights into the reliability of the PD estimate and how it is affected by the described model bias.

\section{Quantifying Uncertainty in PDPs}
\label{sec:uncertainty}

\begin{wrapfigure}[13]{R}{0.5\textwidth}
    \vspace{-2em}
    \centering
    \includegraphics[width = \linewidth]{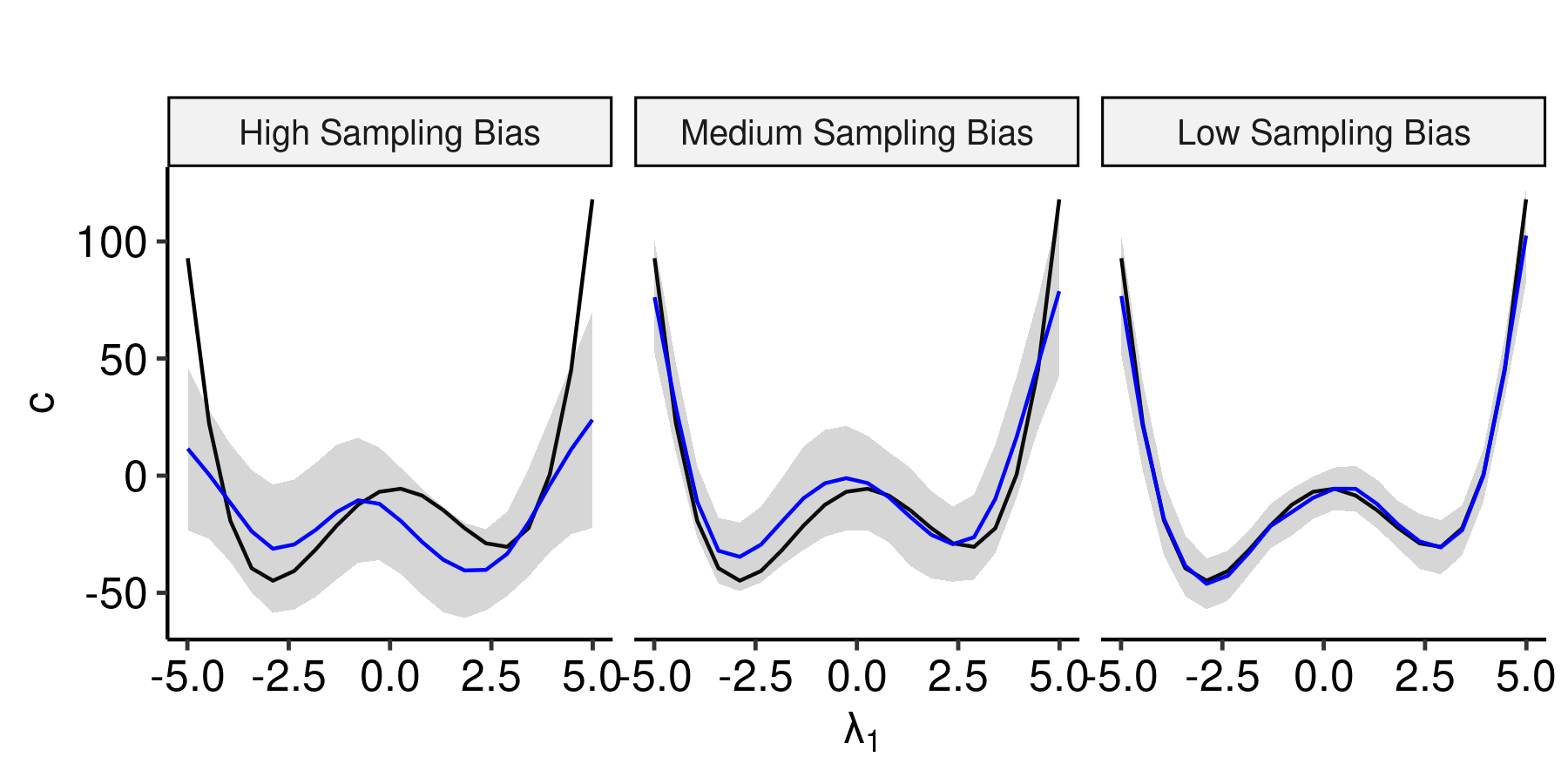}
    \caption{PDPs (blue) with confidence bands for surrogates trained on data created by BO and LCB with $\tau = 0.1$ (left), $\tau = 1$ (middle) and uniform i.i.d. dataset (right) vs. the true PD (black). }
    \label{fig:bias_pdp_test}
    
\end{wrapfigure}

Pointwise uncertainty estimates of a probabilistic model provide insights into the reliability of the prediction $\hat c(\lambdab)$ for a specific configuration $\lambdab$. 
This uncertainty directly correlates with how explored the region around $\lambdab$ is.
Hence, including the model's uncertainty structure into the PD estimate enables users to understand in which regions the PDP is more reliable and which parts of the PDP must be cautiously interpreted.\footnote{Note that we aim at representing model uncertainty in a PD estimate, and not the variability of the mean prediction (see Appendix \ref*{sec:app_choice_uncertainty} for a more detailed justification). }  
We now extend the PDP of Eq. \eqref{eq:estimate_pdp} to probabilistic surrogate models $\hat c$ (e.g., a GP). Let $\lambdab_S$ be a fixed grid point and $\left( \lambdab_C^{(i)}\right)_{i = 1, ..., n} \sim \P\left(\lambdab_C\right)$ a sample that is used to compute the Monte Carlo estimate of Eq. \eqref{eq:estimate_pdp}. 
The vector of predicted performances at the grid point $\lambdab_S$ is $\bm{\hat c}\left(\lambdab_S\right) = \left(\hat c\left(\lambdab_S, \lambdab_C^{(i)}\right)\right)_{i = 1, ..., n}$
with (posterior) mean $\bm{\hat m}\left(\lambdab_S\right) := \left(\hat m\left(\lambdab_S, \lambdab_C^{(i)}\right)\right)_{i = 1, ..., n}$ and a (posterior) covariance $\bm{\hat K}\left(\lambdab_S\right) := \left(\hat k\left(\left(\lambdab_S, \lambdab_C^{(i)}\right), \left(\lambdab_S, \lambdab_C^{(j)}\right)\right)\right)_{i, j = 1, ..., n}$.  
Thus, $\hat c_{S}\left(\lambdab_S\right) = \frac{1}{n} \sum\nolimits_{i = 1}^n \hat c\left(\lambdab_S, \lambdab_C^{(i)}\right)$ is a random variable itself.
The expected value of $\hat c_{S}\left(\lambdab_S\right)$ corresponds to the PD of the posterior mean function $\hat m$ at $\lambdab_S$, i.e.: 
\begin{eqnarray}
    \hat m_S\left(\lambdab_S\right) &=& \E_{\bm{\hat c}} \left[\hat c_{S}\left(\lambdab_S\right)\right] = \E_{\bm{\hat c}} \left[\frac{1}{n}\sum\nolimits_{i = 1}^n \hat c\left(\lambdab_S, \lambdab_C^{(i)}\right)\right] =\frac{1}{n} \sum\nolimits_{i = 1}^n \hat m\left(\lambdab_S, \lambdab_C^{(i)}\right). 
    \label{eq:pdp_expectation}
\end{eqnarray}
The variance of $\hat c_{S}\left(\lambdab_{S}\right)$ is 
\begin{eqnarray}
    \hat s_S^2(\lambdab_S) &=&\mathbb{V}_{\bm{\hat c}}\left[\hat c_{S}\left(\lambdab_S\right)\right] = \mathbb{V}_{\bm{\hat c}} \left[\frac{1}{n} \sum\nolimits_{i = 1}^n \hat c\left(\lambdab_S, \lambdab_C^{(i)}\right) \right] =
   \frac{1}{n^2} \bm{1}^\top \bm{\hat K}\left(\lambdab_S\right) ~ \bm{1}.
    \label{eq:pdp_gp_var}
\end{eqnarray}
For the above estimate, it is important that the kernel is correctly specified such that the covariance structure is modeled properly by the surrogate model. 
Eq. \eqref{eq:pdp_gp_var} can be approximated empirically by treating the pairwise covariances as unknown, i.e.:
\begin{eqnarray}
    \hat s_S^2\left(\lambdab_S\right) &\approx& 
  \frac{1}{n} \sum\nolimits_{i = 1}^n \bm{\hat K}\left(\lambdab_S\right)_{i,i}.
    \label{eq:pdp_gp_var_nocov}
\end{eqnarray}
In Appendix~\ref*{sec:app_covariances}, we show empirically that this approximation is less sensitive to kernel misspecifications. 
Please note that the variance estimate and the mean estimate can also be applied to other probabilistic models, such as GAMLSS\footnote{Generalized additive models for location, scale and shape}, transformation trees, or a random forest.
An example for PDPs with uncertainty estimates is shown in Figure \ref{fig:bias_pdp_test} for different degrees of a sampling bias.







\section{Regional PDPs via Confidence Splitting}
\label{sec:splitting}

\begin{figure}[t]
\begin{minipage}[htb]{0.48\textwidth}
    \centering
    \includegraphics[width = 0.9\linewidth]{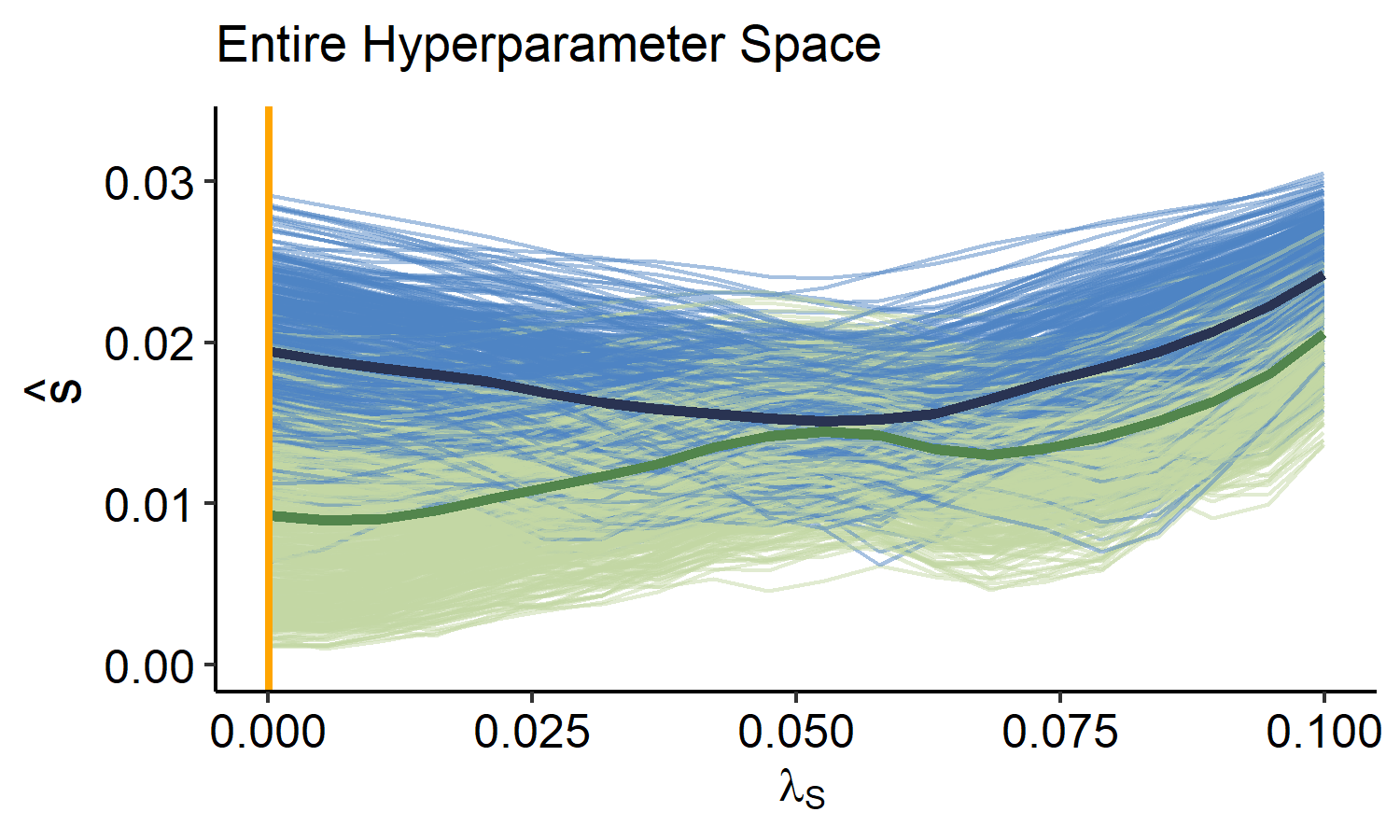}
    \captionof{figure}{ICE curves of $\hat{s}$ of $\lambdab_S$ for the left (green) and right (blue) sub-region after the first split. The darker lines represent the respective PDPs. The orange vertical line marks the value $\lambda_S$ of the optimal configuration.}
    \label{fig:ice_sd}
\end{minipage}\hfill
\begin{minipage}[htb]{0.48\textwidth}
    \scalebox{0.8}{
    \hspace{40pt} 
      \begin{tikzpicture}
      \usetikzlibrary{arrows}
        \usetikzlibrary{shapes}
         \tikzset{treenode/.style={draw, circle, font=\small}}
         \tikzset{line/.style={draw, thick}}
         \node [treenode] (a0) {$\mathcal{N}$};[below=5pt,at=(node1.south)]
         \node [treenode, below=0.4cm, at=(a0.south), xshift=-2.0cm]  (a1) {$\mathcal{N}_l$};
         \node [treenode, below=0.4cm, at=(a0.south), xshift=2.0cm]  (a2) {$\mathcal{N}_r$};
         \path [line] (a0.south) -- + (0,-0.2cm) -| (a1.north) node [midway, above] {$\lambda_j< 6.9$};
         \path [line] (a0.south) -- +(0,-0.2cm) -|  (a2.north) node [midway, above] {$\lambda_j\geq6.9$};
      \end{tikzpicture}
      } \\
  \includegraphics[width = 0.49\linewidth]{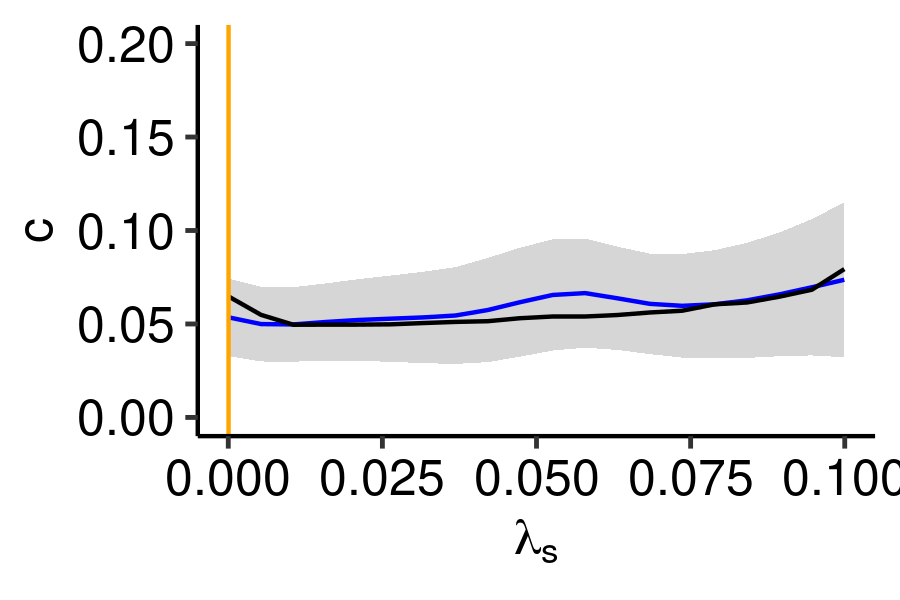}\includegraphics[width = 0.49\linewidth]{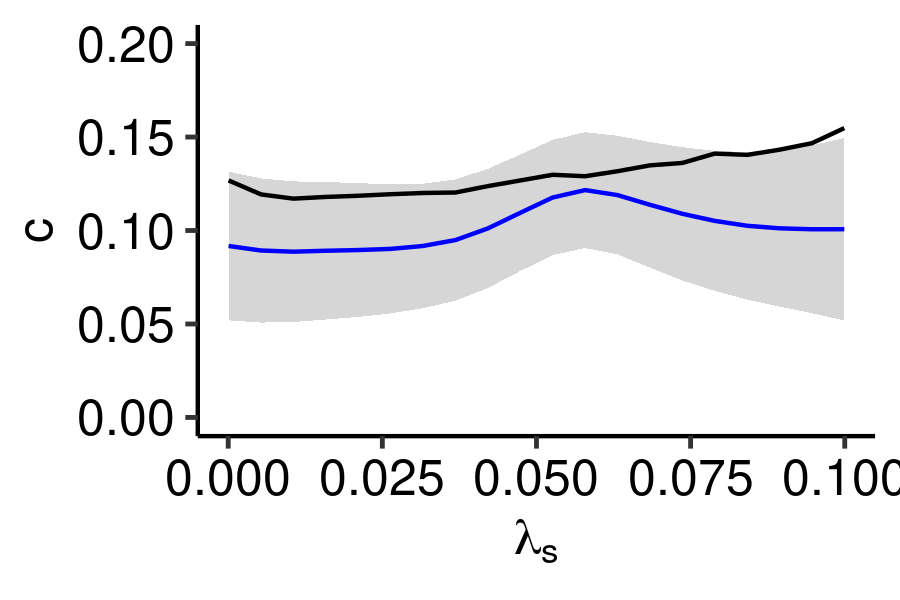}
    \captionof{figure}{Example of two estimated PDPs (blue line) and $95\%$ confidence bands after one partitioning step. The orange vertical line is the value of $\lambdab_S$ from the optimal configuration, the black curve is the true PD estimate $c_S(\lambdab_S)$.}
    \label{fig:pdp_explain}
\end{minipage}
\end{figure}

As discussed in Section \ref{sec:bias}, (efficient) optimization may imply that the sampling is biased, which in turn can produce misleading interpretations when IML is naively applied. 
We now aim to identify sub-regions $\Lambda^\prime \subset \Lambda$ of the hyperparameter space in which the PD can be estimated with high confidence, and separate those from sub-regions in which it cannot be estimated reliably. In particular, we identify sub-regions in which poorly-learned effects do not obfuscate the well-learned effects along each grid point, thereby allowing the user to draw conclusions with higher confidence.
By partitioning the entire hyperparameter space through a tree-based approach into disjoint and interpretable sub-regions, a more detailed understanding of the sampling process and hyperparameter effects is achieved. Users can either study the hyperparameter effect of a (confident) sub-region individually or understand the exploration-exploitation sampling of HPO by considering the complete tree structure.
The result of this procedure for a single split is shown in Figure \ref{fig:pdp_explain}.

The PD estimate on the \emph{entire} hyperparameter space $\Lambda$ is computed by sampling the Monte Carlo estimate $(\lambdab_C^{(i)})_{i \in \mathcal{N}} \sim \P(\lambdab_C)$, $\mathcal{N} := \{1, 2, ..., n\}$. 
We now introduce the PD estimate on a \emph{sub-region} $\Lambda^\prime \subset \Lambda$ simply as $(\lambdab_C^{(i)})_{i \in \mathcal{N^\prime}}$ only using $\mathcal{N}^\prime = \{i \in \mathcal{N}\}_{\lambdab^{(i)} \in \Lambda^\prime}$. Since we are interested in the marginal effect of the hyperparameter(s) $S$ at each $\lambdab_S \in \Lambda_S$, we will usually visualize the PD for the whole range $\Lambda_S$. Thus, all obtained sub-regions should be of the form $\Lambda^\prime = \Lambda_S \times \Lambda_C^\prime$ with $\Lambda_C^\prime\subset \Lambda_C.$ 
This corresponds to an average of ICE curves in the set $i \in \mathcal{N}^\prime$. 
The pseudo-code to partition a hyperparameter (sub-)space $\Lambda$ and corresponding sample  
$(\lambdab_C^{(i)})_{i \in \mathcal{N}} \in \Lambda_C$, $\mathcal{N} \subseteq \{1, ..., n\}$, 
into two child regions is shown in Algorithm~\ref{alg:tree}. 
This splitting is recursively applied in a CART\footnote{Classification and regression trees}-like procedure~\citep{breiman:1984} to expand a full tree structure, with the usual stopping criteria 
(e.g., a maximum number of splits, a minimum size of a region, or a minimum improvement in each node).
In each leaf node, the sub-regional PDP and its corresponding uncertainty estimate are computed by aggregating over all contained ICE curves. 

The criterion to evaluate a specific partitioning is based on the idea of grouping ICE curves with similar uncertainty structure. 
To be more exact, we evaluate the impurity of a PD estimate on a sub-region $\Lambda^\prime$ with the help of the associated set of observations $\mathcal{N}^\prime = \{i \in \mathcal{N}\}_{\lambdab_C^{(i)} \in \Lambda_C^\prime}$, also referred to as nodes, as follows: For each grid point $\lambdab_S$, we use the L2 loss in $ L\left(\lambdab_S, \mathcal{N}^\prime\right)$ to evaluate how the uncertainty varies across all ICE estimates $i \in \mathcal{N}^\prime$ using $\hat s^2_{S|\mathcal{N}^\prime} \left(\lambdab_S\right):= \frac{1}{|\mathcal{N}^\prime|}\sum_{i \in \mathcal{N}^\prime} \hat s^2\left(\lambdab_S, \lambdab_C^{(i)}\right)$ and aggregate the loss $ \mathcal{L}\left(\lambdab_S, \mathcal{N}^\prime\right)$ over all grid points in $\mathcal{R}_{L2}(\mathcal{N}^\prime)$:
\begin{eqnarray}
    \mathcal{L}\left(\lambdab_S, \mathcal{N}^\prime\right) = \sum\nolimits_{i \in \mathcal{N}^\prime}\left(\hat s^2\left(\lambdab_S, \lambdab_C^{(i)}\right) - \hat s^2_{S|\mathcal{N}^\prime} \left(\lambdab_S\right)\right)^2 \text{ and } \mathcal{R}_{L2}(\mathcal{N}^\prime) = \sum\nolimits_{g = 1}^G \mathcal{L} (\lambdab_S^{(g)}, \mathcal{N}^\prime). 
    \label{eq:splitting_crit_L2}
\end{eqnarray}
 \begin{wrapfigure}{l}{0.5\textwidth}
\begin{minipage}[htb]{0.48\textwidth}
  \begin{algorithm}[H]
    \caption{Tree-based Partitioning}
    \label{alg:tree}
    \begin{algorithmic}
      \STATE \textbf{input: } $\mathcal{N}$
      \FOR{$j \in C$}{
        \FOR{Every split $t$ on hyperparameter $\lambda_j$ }{
            \STATE $\mathcal{N}_l^{j, t} = \{ i \in \mathcal{N}\}_{\lambda_j^{(i)} \leq t}$
            \STATE $\mathcal{N}_r^{j, t} = \{i \in \mathcal{N}\}_{\lambda_j^{(i)} > t}$ 
            \STATE $\mathcal{I}(j, t) = \mathcal{R}_{L_2}(\mathcal{N}_l^{j, t}) + \mathcal{R}_{L_2}(\mathcal{N}_r^{j, t})$
        }\ENDFOR
      } \ENDFOR   
        \STATE Choose $\left(j^\ast, t^\ast_{\lambda^\ast_j}\right) \in \argmin\nolimits_{j, t} \mathcal{I}(j, t)$
        \STATE Return $\mathcal{N}_l^{j, t}$ and $\mathcal{N}_r^{j, t}$ for $(j, t) = \left(j^\ast, t^\ast_{\lambda^\ast_j}\right)$
    \end{algorithmic}
  \end{algorithm}
    \centering
    \end{minipage} 
\end{wrapfigure}
Hence, we measure the pointwise $L_2$-distance between ICE curves of the variance function $\hat s^2(\lambdab_S, \lambdab_C^{(i)})$ and its PD estimate $\hat s^2_{S|\mathcal{N}^\prime}\left(\lambdab_S\right)$ within a sub-region $\mathcal{N}^\prime$. 
This seems reasonable, as ICE curves in well-explored regions of the search space should, on average, have a lower uncertainty than those in less-explored regions.
However, since we only split according to hyperparameters in $C$ but not in $S$, 
the partitioning does not cut off less explored regions w.r.t. $\lambdab_S$. 
Thus, the chosen split criterion groups ICE curves of the uncertainty estimate such that we receive sub-regions associated with low costs~$c$ (and thus high relevance for a user) to be more confident in well-explored regions of $\lambdab_S$ and less confident in under-explored regions.  
Figure \ref{fig:ice_sd} shows that ICE curves of the uncertainty measure with high uncertainty over the entire range of $\lambdab_S$ are grouped together (right sub-region). Those with low uncertainty close to the optimal configuration of $\lambdab_S$ and increasing uncertainties for less suitable configurations are grouped together by curve similarities in the left sub-region. 
The respective PDPs are illustrated in Figure \ref{fig:pdp_explain}, where the confidence band in the left sub-region decreased compared to the confidence band of the global PDP especially for grid points close to the optimal value of $\lambdab_S$. Hence, by grouping observations with similar ICE curves of the variance function, resulting sub-regional PDPs with confidence bands provide the user with the information of which sub-regions of $\Lambda_C$ are well-explored and lead to more reliable PDP estimates. Furthermore, the user will know which ranges of $\lambdab_S$ can be interpreted reliably and which ones need to be regarded with caution.


To sum up, the splitting procedure provides interpretable, disjoint sub-regions of the hyperparameter space. Based on the defined impurity measure, PDPs with high reliability can be identified and analyzed. In particular, the method provides more confident and reliable estimates in the sub-region containing the optimal configuration.
Which PDPs are most interesting to explore depends on the question the user would like to answer. If the main interest lies in understanding the optimization and exploring the sampling process, a user might want to keep the number of sub-regions relatively low by performing only a few partitioning steps. Subsequently, one would investigate the overall structure of the sub-regions and the individual sub-regional PDPs. If users are more interested in interpreting hyperparameter effects only in the most relevant sub-region(s), they may want to split deeper and only look at sub-regions that are more confident than the global PDP.

Due to the nature of the splitting procedure, the PDP estimate on the entire hyperparameter space is a weighted average of the respective sub-regional PDPs. Hence, the global PDP estimate is decomposed into several sub-regional PDP estimates. Furthermore, note that the proposed method does not assume a numeric hyperparameter space, since the uncertainty estimates as well as ICE and PDP estimates can also be calculated for categorical features. Thus, it is applicable to problems with mixed spaces as long as a probabilistic surrogate model -- and particularly its uncertainty estimates -- are available. In Appendix \ref*{sec:app_hierarchical} we describe how our method is applied to hierarchical hyperparameter spaces. 


Since the proposed method is an instance of the CART algorithm, finding the optimal split for a categorical variable with $q$ levels generally involves checking $2^q$ subsets.
This becomes computationally infeasible for high values of $q$. 
It remains an open question for future work if this can be sped by an optimal procedure as in regression with L2 loss \citep{fisher1958grouping} and binary classification \citep{breiman1984trees} or by a clever heuristic as for multiclass classification \citet{wright2019splitting}.

\section{Experimental Analysis}
\label{sec:experiments}

In this section, we validate the effectiveness of the introduced methods.
We formulate two main hypotheses: First, experimental data affected by the sampling bias lead to biased surrogate models and thus to unreliable and misleading PDPs.
Second, the proposed partitioning allows us to identify an interpretable sub-region (around the optimal configuration) that yields a more reliable and confident PDP estimate. In a first experiment, we apply our methods to BO runs on a synthetic function. In this controlled setup, we investigate the validity of our hypotheses with regards to problems of different dimensionality and different degrees of sampling bias.
In a second experiment, we evaluate our PDP partitioning in the context of HPO for neural networks on a variety of tabular datasets.

We assess the sampling bias of the optimization design points by comparing their empirical distribution to a uniform distribution via Maximum Mean Discrepancy (MMD) \citep{gretton:2012, molnar2020modelagnostic}, which is covered in more detail in the Appendix~\ref*{sec:app_experimental_metrics}. 
We measure the reliability of a PDP, i.e., the degree to which a user can rely on the estimate of the PD estimate, by comparing it to the true PD $c_S(\lambdab_S)$ as defined in Eq. \eqref{eq:pdp}. More specifically, for every grid point $\lambdab_S^{(g)}$, we compute the negative log-likelihood (NLL) of $c_S(\lambdab_S)$ under the distribution of $\hat c_S\left(\lambdab_S\right)$ pointwise for every grid point $\lambdab_S^{(g)}$. 
The confidence of a PDP is illustrated by the width of its confidence bands $\hat m_S\left(\lambdab_S\right) \pm q_{1 - \alpha/2}\cdot\hat s_S\left(\lambdab_S\right)$, with $q_{1 - \alpha/2}$ denoting the $(1 - \alpha / 2)$-quantile of a standard normal distribution. We measure the confidence by assessing $\hat s_S(\lambdab_S)$ pointwise for every grid point. In particular, we consider the mean confidence (MC) across all grid points $\frac{1}{G} \sum_{g = 1}^G \hat s\left(\lambdab_S^{(g)}\right)$ as well as the confidence at the grid point closest to $\hat{\lambdab}_S$ abbreviated by OC, with $\hat{\lambdab}$ being the best configuration evaluated by the optimizer. 
To evaluate the performance of the confidence splitting, we report the above metrics on the sub-region that contains the best configuration evaluated by the optimizer, assuming that this region is of particular interest for a user of HPO.
PDPs are computed with regards to single features for $G = 20$ equidistant grid points and $n = 1000$ Monte Carlo samples.

\subsection{BO on a Synthetic Function}
\label{sec:hpo_synth}
We consider the $d$-dimensional Styblinski-Tang function $c: \left[-5, 5\right]^d \to \R$, 
$\lambdab \mapsto \frac{1}{2} \sum_{i = 1}^d \left(\lambdab_i^4 + 16 \lambdab_i^2 + 5 \lambdab_i\right)$ for $d \in \{3, 5, 8\}$. 
Since the PD is the same for each dimension $i$, we only present the effects of $\lambdab_1$.
We performed BO with a GP surrogate model with a Mat\'{e}rn-3/2 kernel and the LCB acquisition function $a(\lambdab) = \hat m(\lambdab) + \tau \cdot \hat s(\lambdab)$ with different values $\tau \in \{0.1, 1, 5\}$ to control the sampling bias. We compute the global PDP with confidence bands estimated according to Eq. \eqref{eq:pdp_gp_var_nocov} for the GP surrogate model $\hat c$ that was fitted in the \emph{last} iteration of BO. We ran Algorithm~\ref{alg:tree}, and computed the PDP in the sub-region containing the optimal configuration. All computations were repeated $30$ times. Further details on the setup are given in Appendix~\ref*{sec:app_experimental_design_synth}.

\begin{minipage}[t]{0.4\textwidth}
    \includegraphics[width = \linewidth]{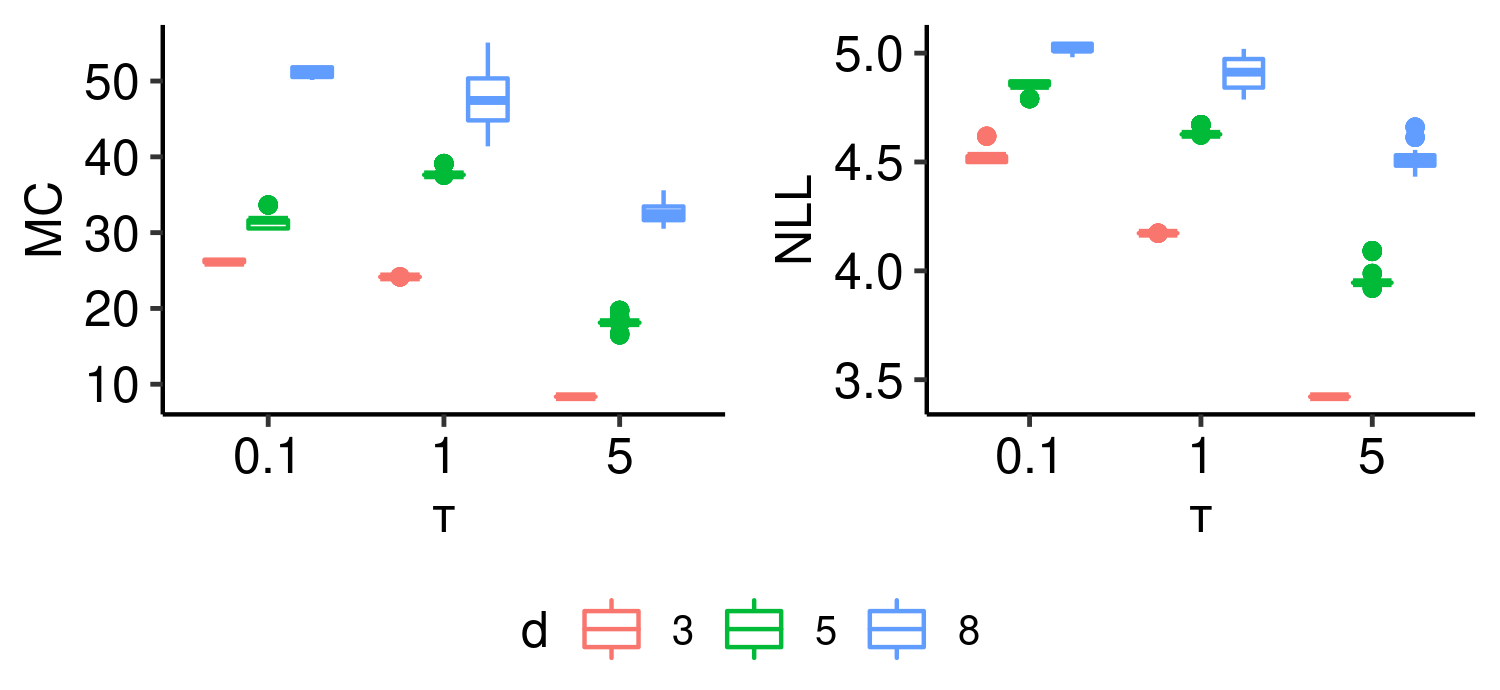}
    \captionof{figure}{The figure presents the MC (left) and the NLL (right) for $d \in \{3, 5, 8\}$ for a high ($\tau = 0.1$), medium ($\tau = 1$), and low ($\tau = 5$) sampling bias across $30$ replications. With a lower sampling bias, we obtain narrower confidence bands and a lower NLL.}
    \label{fig:reliability_and_confidence}
\end{minipage}\hfill
\begin{minipage}[htb]{0.56\textwidth}
 \begin{scriptsize}
\captionof{table}{The table shows the relative improvement of the MC and the NLL via Algorithm \ref{alg:tree} with $1$ and $3$ splits, compared to the global PDP along with the sampling bias for a $\tau = 0.1$ (high), $\tau = 2$ (medium), and $\tau = 5$ (low). Results are averaged across $30$ replications.}\label{tab:tree_depth}
\begin{tabular}{cccccc}
\toprule 
&  & \multicolumn{2}{c}{$\delta$ MC (\%)} &  \multicolumn{2}{c}{$\delta$ NLL (\%)}  \\ \cmidrule{3-4} \cmidrule{5-6} 
$d$ & MMD & $n_\textrm{sp}=1$ & $n_\textrm{sp}=3$ & $n_\textrm{sp}=1$ & $n_\textrm{sp}=3$\\ \midrule
 3 & low (0.18) & 7.65 & 13.64 & 5.89 & 10.92 \\ 
  3 & medium (0.51) & 12.86 & 36.92 & 4.78 & 7.70 \\ 
   3 & high (0.56) & 16.52 & 34.84 & 2.77 & -1.62 \\ 
   5 & low (0.15) & 6.63 & 15.45 & 2.82 & 6.05 \\ 
  5 & medium (0.45) & 19.67 & 37.28 & 4.05 & 7.80 \\ 
   5 & high (0.53) & 11.99 & 33.06 & -3.86 & -1.93 \\ 
   8 & low (0.11) & 3.58 & 9.67 & 0.84 & 2.40 \\ 
  8 & medium (0.42) & 8.86 & 23.03 & 1.51 & 3.30 \\ 
 8 & high (0.56) & 6.59 & 19.84 & 1.53 & 4.29 \\ 
\bottomrule                       
\end{tabular}
    \end{scriptsize}
 \end{minipage} \hfill

As presented in Figure \ref{fig:reliability_and_confidence}, the PDPs for surrogate models trained on \emph{less biased} data (measured by the MMD) yield \emph{lower} values of the NLL, as well as \emph{lower} values for the MC.
Table \ref{tab:tree_depth} shows that a single tree-based split reduces the MC by up to almost $20 \%$, and up to $37\%$ when performing $3$ partitioning steps. Additionally, the NLL improves with an increasing number of partitioning steps in most cases. The results on the synthetic functions support our second hypothesis that the tree-based partitioning improves the reliability in terms of the NLL and the confidence of the PD estimates. The improvement of the MC is higher for a medium to high sampling bias, compared to scenarios that are less affected by sampling bias. We observe that (particularly for high sampling bias) there are some outlier cases in which the NLL worsens. More detailed results are shown in Appendix~\ref*{sec:app_experimental_results_synth}.

\subsection{HPO on Deep Learning}
\label{sec:hpo_nn}

In a second experiment, we investigate HPO in the context of a surrogate benchmark \citep{eggensperger2015efficient} based on the LCBench data \citep{Zimmer2020AutoPyTorchTM}. 
For each of the $35$ different OpenML \citep{vanschoren2014openml} classification tasks, LCBench provides access to evaluations of a deep neural network on $2000$ configurations randomly drawn from the configuration space defined by Auto-PyTorch Tabular (see Table \ref*{tab:searchspace} in Appendix \ref*{sec:app_experimental_design}). For each task, we trained a random forest as an empirical performance model that predicts the balanced validation error of the neural network for a given configuration. 
These empirical performance models serve as cheap to evaluate objective functions, which efficiently approximate the result of the real-world experiment of running a deep learning configuration on an LCBench instance. 
BO then acts on this empirical performance model as its objective\footnote{Please note that the random forest is only used as a surrogate in order to construct an efficient benchmark objective, and not as a surrogate in the BO algorithm, where we use a GP.}. 

For each task, we ran BO to obtain the optimal architecture and hyperparameter configuration.
Again, we used a GP with a Mat\'{e}rn-3/2 kernel and LCB with $\tau = 1$. Each BO run was allotted a budget of $200$ objective function evaluations. We computed the PDPs and their confidences, which are estimated according to Eq. \eqref{eq:pdp_gp_var_nocov}, based on the surrogate model $\hat c$ after the final iteration. We performed tree-based partitioning with up to $6$ splits based on a uniformly distributed dataset of size $n = 1000$. All computations were statistically repeated 30 times. Further details are provided in Appendix \ref*{sec:app_experimental_design_mlp}.

\begin{wraptable}{R}{0.5\textwidth}
        \caption{Relative improvement of MC, OC, and NLL on hyperparameter level. The table shows the respective mean (standard deviation) of the average relative improvement over 30 replications for each dataset and 6 splits.}
        \label{tab:conf.impr.feat}
        \begin{scriptsize}
        \begin{tabular}{lrrrr}
         \toprule
        Hyperparameter & $\delta$ MC (\%) & $\delta$ OC (\%) & $\delta$ NLL (\%) \\ 
         \midrule
        Batch size & 40.8 (14.9) & 61.9 (13.5) & 19.8 (19.5)\\ 
        Learning rate & 50.2 (13.7) & 57.6 (14.4) & 17.9 (20.5)\\ 
        Max. dropout & 49.7 (15.4) & 62.4 (11.9)  & 17.4 (18.2)\\ 
        Max. units & 51.1 (15.2) & 58.6 (12.7) & 24.6 (22.0)\\ 
        Momentum & 51.7 (14.5) & 58.3 (12.7)  & 19.7 (21.7)\\ 
        Number of layers & 30.6 (16.4) & 50.9 (16.6) & 13.8 (32.5)\\ 
        Weight decay & 36.3 (22.6) & 61.0 (13.1) & 11.9 (19.7)\\ 
        \bottomrule
        \end{tabular}
        \end{scriptsize}
\end{wraptable}

For the real-world data example, we focus on answering the second hypothesis, i.e., whether the tree-based Algorithm~\ref{alg:tree} improves the reliability of the PD estimates. We compare the PDP in sub-regions 
after $6$ splits with the global PDP. We computed the relative improvement of the confidence (MC and OC) and the NLL of the sub-regional PDP compared to the respective estimates for the global PDP. As shown in Table \ref{tab:conf.impr.feat}, the MC of the PDPs is on average reduced by $30\%$ to $52\%$, depending on the hyperparameter. At the optimal configuration $\hat{\lambdab}_S$, the improvement even increases to $50\% - 62\%$. Thus, PDP estimates for all hyperparameters are on average -- independent of the underlying dataset -- clearly more confident in the relevant sub-regions when compared to the global PD estimates, especially around the optimal configuration $\hat{\lambdab}_S$. 
In addition to the MC, the NLL simultaneously improves. 
In Appendix \ref*{sec:app_experimental_results_mlp}, we provide details regarding the evaluated metrics on the level of the dataset and demonstrate that our split criterion outperforms other impurity measures regarding MC and~OC. Furthermore, we emphasize in Appendix \ref*{sec:app_experimental_results_mlp} the significance of our results by providing a comparison to a naive baseline method.
\begin{figure}[bth]
    \centering
    \includegraphics[width = 0.9\linewidth]{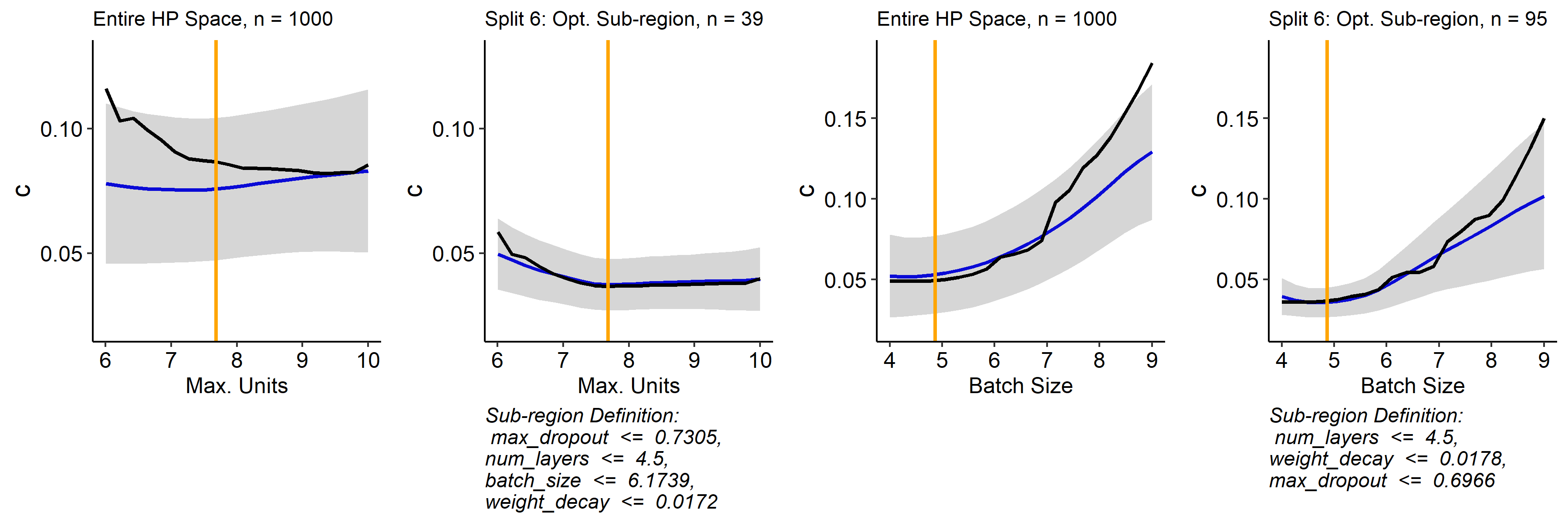}
    \caption{PDP (blue) and confidence band (grey) of the GP for hyperparameter \textit{max. number of units} (\textit{batch size}) on the left (right) side. The black line shows the PDP of the meta surrogate model representing the true PDP estimate. The orange vertical line marks the optimal configuration $\hat{\lambdab}_S$. The relative improvements from the global PDP to the sub-regional PDP 
    after 6 splits are for \textit{max. number of units} (\textit{batch size}): $\delta$ MC = $61.6 \%$ ($28.4 \%$), $\delta$ OC = $63.5 \%$ ($62.2 \%$), $\delta$ NLL = $48.6 \%$ ($30.1 \%$).}
    \label{fig:mlp_example1}
\end{figure}

To further study our suggested method, we now highlight a few individual experiments. We chose one iteration of the \textit{shuttle} dataset. On the two left plots of Figure \ref{fig:mlp_example1}, we see that the true PDP estimate
for \textit{max. number of units} is decreasing, while the globally estimated PDP trend is increasing and thus misleading.
Although the confidence band already indicates that the PDP cannot be reliably interpreted on the entire hyperparameter space, it remains challenging to draw any conclusions from it. After performing $6$ splits, we receive a confident and reliable PD estimate on an interpretable sub-region. 
The same plots are depicted for the hyperparameter \textit{batch size} on the right part of Figure \ref{fig:mlp_example1}. This example illustrates that the confidence band might not always shrink uniformly over the entire range of $\lambdab_S$ during the partitioning, but often particularly around the optimal configuration $\hat{\lambdab}_S$. 

\section{Discussion and Conclusion}
\label{sec:conclusion}
In this paper, we showed that partial dependence estimates for surrogate models fitted on experimental data generated by efficient hyperparameter optimization can be unreliable due to an underlying sampling bias.
We extended PDPs by an uncertainty estimate to provide users with more information regarding the reliability of the mean estimator. 
Furthermore, we introduced a tree-based partitioning approach for PDPs, where we leverage the uncertainty estimator to decompose the hyperparameter space into interpretable, disjoint sub-regions. 
We showed with two experimental studies that we generate, on average, more confident and more reliable regional PDP estimates in the sub-region containing the optimal configuration compared to the global PDP.

One of the main limitations of PDPs is that they bear the risk of providing misleading results if applied to correlated data in the presence of interactions, especially for nonparametric models \citep{groemping:2020}. 
However, existing alternatives that visualize the global marginal effect of a feature such as accumulated local effect (ALE) plots \citep{apley2020visualizing} do also not provide a fully satisfying solution to this problem \citep{groemping:2020}. As a solution to this problem, 
\cite{groemping:2020} suggests stratified PDPs by conditioning on a correlated and potentially interacting feature to group ICE curves. 
This idea is in the spirit of our introduced tree-based partitioning algorithm. 
However, in the context of BO we might assume the distribution in Eq.~\eqref{eq:pdp} to be uniform and therefore no correlations are present. 
Instead of correlated features, we are faced with a sampling bias (see Section~\ref{sec:bias}) where we observe regions of varying uncertainty. 
Hence, instead of stratifying with respect to correlated features and aggregating ICE curves in regions with less correlated features, we stratify with respect to uncertainty and aggregate ICE curves in regions with low uncertainty variation. 
Nonetheless, it might be interesting to compare our approach with approaches based on the considerations made by \cite{groemping:2020} -- or potentially improved ALE curves.

Another limitation when using single-feature PDPs as in our examples is that hyperparameter interactions are not visible. However, two-way interactions can be visualized by plotting two-dimensional PDPs within sub-regions. Another possibility to detect interactions is to look at ICE curves within the sub-regions.
If the shape of ICE curves within a sub-region is very heterogeneous, it indicates that the hyperparameter under consideration interacts with one of the other hyperparameters. 
Hence, having the additional possibility to look at ICE curves of individual observations within a sub-region is an advantage compared to other global feature effect plots such as ALE plots \citep{apley2020visualizing}, as they are not defined on an observational level.
While we mainly discussed GP surrogate models on a numerical hyperparameter space in our examples, our methods are applicable to a wide variety of distributional regression models and also for mixed and hierarchical hyperparameter spaces. 
We also considered in Appendix~\ref*{sec:app_experimental_results_mlp} different impurity measures. While the one introduced in this paper performed best in our experimental settings, this impurity measure as well as other components are exchangeable within the proposed algorithm. 
In the future, we will study our method on more complex, hierarchical configuration spaces for neural architecture search. 

The proposed interpretation method is based on a surrogate and consequently does provide insights about what the AutoML system has \emph{learned}, which in turn allows plausibility checks and may increase trust in the system. To what extent this allows conclusions on the \emph{true} underlying hyperparameter effects depends on the quality of the surrogate. How to efficiently perform model diagnostics to ensure a high surrogate quality before applying interpretability techniques is subject to future research. 

While we focused on providing better explanations without generating any additional experimental data, it might be interesting to investigate in future work how confidence and reliability of IML methods can be increased most efficiently when a user is allowed to conduct additional experiments. 

Overall, we believe that increasing interpretability of AutoML will pave the way for human-centered AutoML. Our vision is that users will be able to better understand the reasoning and the sampling process of AutoML systems and thus can either trust and accept the results of the AutoML system or interact with it in a feedback loop based on the gained insights and their preferences. How users can then best interact with AutoML (beyond simple changes of the configuration space) will be left open for future research.





 \begin{ack}


This work has been partially supported by the German Federal Ministry of Education and Research (BMBF) under Grant No. 01IS18036A. The authors of this work take full responsibilities for its content.
\end{ack}

\newpage

\small

\bibliography{Bib}


\newpage

\appendix
\hypersetup{bookmarksdepth=-2}

\begin{NoHyper}
\section{Uncertainty Estimation}

\normalsize

\subsection{Choice of Uncertainty Quantification}
\label{sec:app_choice_uncertainty}
Besides using the uncertainty estimate of the surrogate model to quantify the uncertainty for the PDP mean estimate (our method), it is also possible to estimate uncertainty w.r.t. the variance over different ICE curves. However, if the uncertainty was estimated via computing the variance over ICE curves, we describe how the \emph{levels of the mean prediction vary} along the $\boldsymbol{\lambda}_S$ dimensions. In contrast, we propose to capture \emph{model uncertainty} along the $\boldsymbol{\lambda}_S$ dimensions. For example, consider a constant surrogate function $\hat c(\boldsymbol{\lambda}) = \gamma$ with high uncertainty estimation $\hat s^2(\boldsymbol{\lambda}) = 100$. Computing the variance over ICE curves on this example will result in an uncertainty estimate of $0$ (all ICE curves are identical). Our method, however, would return a variance estimate of $100$ and thus capture model uncertainty.

\subsection{Covariance Estimates under Misspecification of Kernels}
\label{sec:app_covariances}

\begin{figure*}[htp]
    \centering
    \includegraphics[width=0.9\textwidth]{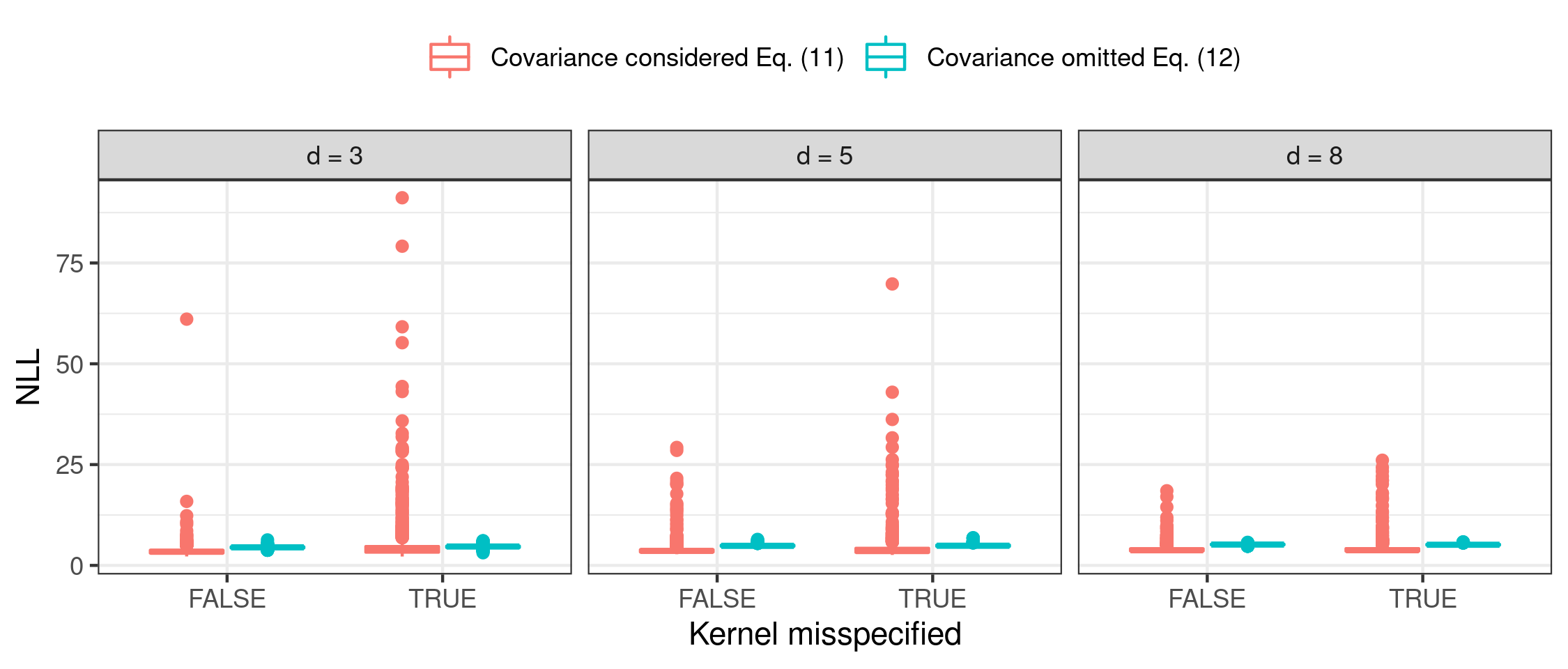}
    \caption{The figures show the NLL of the true PDP $c_1(\lambdab_1)$ under the estimated PDPs with variance estimates \eqref{eq:pdp_gp_var} and \eqref{eq:pdp_gp_var_nocov} and for a misspecified kernel (Gaussian) and a correctly specified kernel (Mat\'{e}rn-3/2), respectively.}
    \label{fig:kernel_misspec}
\end{figure*}

\begin{table*}[ht]
\caption{The table shows the NLL of the true PDP $c_1(\lambdab_1)$ under the estimated PDPs with variance estimates \eqref{eq:pdp_gp_var} and \eqref{eq:pdp_gp_var_nocov} and for a misspecified kernel (Gaussian) and a correctly specified kernel (Mat\'{e}rn-3/2), respectively. Shown are the mean across $50$ replications, and the standard deviation in brackets.}
\label{tab:kernel_misspec}
\centering
\begin{tabular}{rlllll}
\toprule 
&  \multicolumn{2}{c}{Correct specification} & & \multicolumn{2}{c}{Misspecification}  \\ \cmidrule{2-3} \cmidrule{5-6} 
d & Estimate \eqref{eq:pdp_gp_var_nocov} & Estimate \eqref{eq:pdp_gp_var} && Estimate \eqref{eq:pdp_gp_var_nocov} & Estimate \eqref{eq:pdp_gp_var} \\ 
  \midrule
3 & 3.61 (2.02) & 4.47 (0.27) && 5.10 (5.91) & 4.62 (0.32) \\ 
  5 & 3.93 (2.00) & 4.87 (0.23) && 4.33 (3.72) & 4.89 (0.28) \\ 
  8 & 4.05 (1.12) & 5.18 (0.14) && 4.24 (2.12) & 5.13 (0.17) \\ 
   \bottomrule
\end{tabular}
\end{table*}

In order to provide evidence for the claim that estimate Eq.~\eqref{eq:pdp_gp_var} is more sensitive to misspecifications in the kernel (and thus in the covariance structure) than Eq.~\eqref{eq:pdp_gp_var_nocov}, we performed some prior experiments. 

We assume that we are given an objective function that is generated by a Gaussian process (GP) with a Mat\'{e}rn-3/2 kernel. In our experiments, that function was created by fitting a GP on tuples $\left(\lambdab^{(i)}, y^{(i)}\right)_{i = 1, ..., 30}$, with $\lambdab^{(i)} \sim \textrm{Unif}\left([-5, 5]^d\right)$ and $y^{(i)}$ corresponding to the value of the $d$-dimensional Styblinski Tang function for $\lambdab^{(i)}$. The posterior mean of this GP will further serve as our true objective $c$ to pretend that we know the correct kernel specification of the ground-truth. Subsequently, we fit both a GP surrogate model with correctly specified kernel (i.e., a Mat\'{e}rn-3/2 kernel) and a surrogate model with a misspecified kernel (in our case, we chose a Gaussian kernel) to the data $\left(\lambdab^{(i)}, c\left(\lambdab^{(i)}\right)\right)_{i = 1, ..., 30}$. In both cases, we compute the PDPs for $\lambdab_1$ with both variance estimates Eq.~\eqref{eq:pdp_gp_var} and Eq.~\eqref{eq:pdp_gp_var_nocov} and measure the negative log-likelihood (NLL) of $c_S$ under the respective estimated PDP. We performed $50$ repetitions of the experiments for $d \in \{3, 5, 8\}$, respectively. 

Figure \ref{fig:kernel_misspec} shows that the median of the NLL across all $50$ replications is \emph{slightly} lower for the covariance estimate in Eq.~\eqref{eq:pdp_gp_var}. However, the variance of the NLL is much higher for estimate in Eq.~\eqref{eq:pdp_gp_var} as compared to Eq.~\eqref{eq:pdp_gp_var_nocov}. Table \ref{tab:kernel_misspec} confirms that, when using variance estimate Eq.~\eqref{eq:pdp_gp_var}, the standard deviation of the NLL values is lower. We conclude that the reliability of the estimate is particularly sensitive to a correct choice of the kernel function. The NLL for the PDPs computed with variance estimate Eq.~\eqref{eq:pdp_gp_var_nocov} is - independent of whether the kernel is correctly specified or not - less sensitive to misspecifications in the kernel.

\section{Hierarchical Hyperparameter Spaces}
\label{sec:app_hierarchical}
Search spaces in HPO and AutoML are often hierarchical, i.e., some hyperparameters are only active conditional on the value of another hyperparameter (the latter usually being a categorical choice, e.g., using a certain method).
The underlying dependency structure can be visualized by a tree structure (Figure \ref{fig:dependencies} (a)). If one hyperparameter can activate another hyperparameter, we call the former a \emph{parent} and the latter is \emph{subordinate} to the former and called a \emph{child}. A hyperparameter that has no parents is called a global hyperparameter.
Sampled configurations can be presented in a nested block matrix (Figure \ref{fig:dependencies} (b)), with missing entries for inactive hyperparameters. 

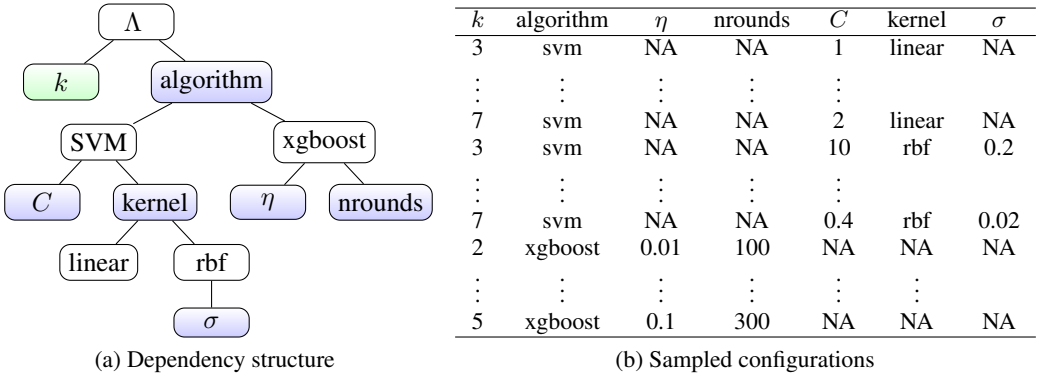
\begin{figure}[h]
    \centering
    \subfloat[Dependency structure]{    
    \begin{tikzpicture}[sibling distance=10em, level distance = 0.8cm,
      every node/.style = {shape=rectangle, rounded corners,
        draw, align=center,
        top color=white, minimum width=1cm}]]
        \tikzstyle{level 1}=[sibling distance=20mm] 
        \tikzstyle{level 2}=[sibling distance=30mm] 
        \tikzstyle{level 3}=[sibling distance=15mm] 
    
      \node { $\Lambda$ }
        child { node[bottom color=green!20] {$k$} }
        child { node[bottom color=blue!20] {algorithm} 
            child { node {SVM} 
                child { node[bottom color=blue!20] {$C$}   }
                child { node[bottom color=blue!20] {kernel} 
                    child { node {linear} }
                    child { node {rbf}
                        child { node[bottom color=blue!20] {$\sigma$} } }
                }
            }
            child { node {xgboost} 
                child { node[bottom color=blue!20] {$\eta$} }
                child { node[bottom color=blue!20] {nrounds} }
            }
        };
    \end{tikzpicture}}\hfill
    \subfloat[Sampled configurations]{
    \begin{footnotesize}
    \begin{tabular}[b]{ ccccccc } 
        \hline
        $k$ & algorithm & $\eta$ & nrounds & $C$ & kernel & $\sigma$\\
         \hline
        3 & svm & NA & NA & 1 & linear & NA \\ 
        \vdots & \vdots & \vdots & \vdots & \vdots\\
        7 & svm & NA & NA & 2 & linear & NA \\ 
        3 & svm & NA & NA & 10 & rbf & 0.2 \\ 
        \vdots & \vdots & \vdots & \vdots & \vdots\\
        7 & svm & NA & NA & 0.4 & rbf & 0.02 \\ 
        2 & xgboost & 0.01 & 100 & NA & NA & NA \\ 
         \vdots & \vdots & \vdots & \vdots & \vdots & \vdots \\
        5 & xgboost & 0.1 & 300 & NA & NA & NA\\ 
         \hline
    \end{tabular}    
    \end{footnotesize}
    }
  \caption{Hyperparameter $k \in \N$ represents the number of selected components of PCA, applied as a preprocessing step. While it is global, always active, and has no subordinate children, 
 \textit{algorithm} -- dependent on its values \textit{SVM} and \textit{xgboost} -- is parent to hyperparameters $C$, \textit{kernel}, $\eta$ and \textit{nrounds}.}
  \label{fig:dependencies}
\end{figure}

Various surrogate models exist, which can learn on such hierarchical spaces, e.g., GPs with specialized kernels \citep{LevesqueDGS17, Swersky2014RaidersOT}, specialized trees in random forests \citep{HutXuHooLey14} or handling the missing entries through imputation \citep{bischl2018mlrmbo}.


The specialized tree algorithm (in normal AutoML, without PDP) can be briefly summarized as follows: We run a normal recursive partitioning as in CART, but in each split, a hyperparameter is only then eligible for potential splitting if the path leading to the current node satisfies all of its preconditions and therefore activates it. 

In order to generalize this tree building in the context of hierarchical dependency structures to a regional PDP computation as in Section \ref{sec:splitting}, we now make the following three modifications:

First, after having selected a hyperparameter $\lambdab_S$ for which we want to estimate the PDP, we subset the uniformly sampled test data on which we are going to fit our regional PDP algorithm to all rows for which $\lambdab_S$ is not missing, i.e., only to configurations in which $\lambdab_S$ is active\footnote{This will likely result in constant values for hyperparameters in the path leading to $\lambdab_S$, and consequently the tree will not split on these.}. 

Second, we now run the specialized tree algorithm as described above, but never allow to split on $\lambdab_S$. Obviously, this will then never activate any child of $\lambdab_S$, so we can never split on these.

Third, we now adapt the estimation of the PDPs in a given tree node and the associated data block $\mathcal{N}$
\begin{equation}
    \hat c_S\left(\lambdab_S\right) = \frac{1}{|\mathcal{N}|} \sum\nolimits_{i \in \mathcal{N}} \hat m\left(\lambdab_S, \lambdab_{C}^{(i)}\right)
    \qquad
    \hat s^2_S\left(\lambdab_S\right) = \frac{1}{|\mathcal{N}|} \sum\nolimits_{i \in \mathcal{N}} \hat s^2\left(\lambdab_S, \lambdab_{C}^{(i)}\right)
    \label{eq:estimate_pdp2}
\end{equation}
in the presence of hierarchical structures. 
If hyperparameter $\lambdab_S$ is a parent, 
sampling from the marginal $\lambdab_C \sim \P(\lambdab_C)$ (which refers to all hyperparameters except $\lambdab_S$) can still yield invalid combinations $\left(\lambdab_S, \lambdab_C\right)$.

We now simply average only over valid configurations w.r.t. to the dependency structure. Formally, let $q(\lambdab): \Lambda \longrightarrow \{0,1\}$ be a binary predicate function which is 1 if and only if $\lambdab$ is valid w.r.t. to the dependency structure.
Now let $v(\lambdab_S, \mathcal{N}) = \{i \in \mathcal{N} | q(\lambdab_S, \lambdab_{C}^{(i)}) = 1\} \subset \mathcal{N}$ be the set of valid configurations in $\mathcal{N}$ w.r.t. the dependency structure if we insert a given $\lambdab_S$ value.
The dependency adapted PDPs now are:
\begin{equation}
    \hat c_S\left(\lambdab_S\right) = \frac{1}{|v(\lambdab_S, \mathcal{N})|} \sum\limits_{i \in v(\lambdab_S, \mathcal{N})} \hat m\left(\lambdab_S, \lambdab_{C}^{(i)}\right)
    \label{eq:estimate_pdp3}
\end{equation}
with an analogous modification for $\hat s^2_S\left(\lambdab_S\right)$.

For example, to calculate the PDP of a child hyperparameter such as \texttt{nrounds} w.r.t. the example in Figure \ref{fig:dependencies}, we first need to subset the test dataset to all rows for which \texttt{nrounds} is not missing (in this specific example this is the same as only keeping the instances where \texttt{algorithm} takes the value \texttt{xgboost}). Due to the dependency structure only $k$, \texttt{algorithm} and $\eta$ are active hyperparameters in $\lambdab_{C}$. 
To calculate the PDP of a parent hyperparameter such as \texttt{algorithm}, there are (in this example) no missings w.r.t. $\lambdab_S$.
However, we will obtain invalid configurations, e.g., when we replace the value \texttt{svm} by \texttt{xgboost} for the parent hyperparameter \texttt{algorithm}. Thus, we need to use the third modification from the above described adjustments and average only over all valid configurations w.r.t. to the dependency structure: if we insert \texttt{svm}, then the invalid configurations are dropped and we average only over those configurations that were activated when choosing \texttt{svm}, i.e., that contain non missing values in the child hyperparameters of the \texttt{svm} algorithm. If we insert \texttt{xgboost} then we only average over those configurations that contain non missing values for $\eta$ and \texttt{nrounds}. 

\section{Experimental Analysis}
\label{sec:app_experimental}

\subsection{Maximum Mean Discrepancy}
\label{sec:app_experimental_metrics}

In Section \ref{sec:hpo_synth} we analyze the first hypothesis how the sampling bias affects the PDP estimation. An indicator of the size of the sampling bias is the exploration factor $\tau$. The smaller $\tau$ the higher the sampling bias compared to a uniformly distributed dataset (e.g. see Figure \ref{fig:sampling_bias}). To put it in other words, the sampling bias can be quantified by the distributional shift between a uniformly distributed sample and the sample generated by the BO process. A commonly used measure to quantify such distributional differences is the Kullback-Leibler divergence. However, since the joint distribution of the generated sample is not known, the Kullback-Leibler divergence might not be the most appropriate measure here.  
Another metric that tests if two different samples belong to the same distribution, is the maximum mean discrepancy (MMD) \citep{gretton:2012}. It is defined by the maximum deviation in expectation and based on the function class of reproducing kernel Hilbert space (RKHS). This is equivalent to `the norm of the difference between distribution feature means in the RKHS' \citep{gretton:2012}.

An unbiased empirical estimate of the MMD for samples $X = \{\bm{x}^{(1)}, ..., \bm{x}^{(n)}\}$ and $Y = \{\bm{y}^{(1)}, ..., \bm{y}^{(m)}\}$ is given by

\begin{align*}
\text{MMD}^2(X,Y) ~=~ & \frac{1}{n(n-1)} \sum_{i=1}^n \sum_{j \neq i}^n k\left(\bm{x}^{(i)},\bm{x}^{(j)}\right)  + \frac{1}{m(m-1)} \sum_{i=1}^m \sum_{j \neq i}^m k\left(\bm{y}^{(i)}, \bm{y}^{(j)}\right) \\ & - \frac{2}{nm} \sum_{i=1}^n \sum_{j=1}^m k\left(\bm{x}^{(i)}, \bm{y}^{(j)}\right)   
\end{align*}

It follows that for $X$ and $Y$ being drawn from the same distribution, the MMD is small while it becomes large for increasing distributional differences.

Here, $X$ represents the sample that is drawn from a uniform distribution over the hyperparameter space $\Lambda$, while $Y$ is the sample generated by the BO process. The kernel $k$ is chosen to be the radial basis function kernel with parameter $\sigma$ being set to the median L2-distance between sample points. This heuristic is commonly used \citep{gretton:2012}.

\subsection{Experimental Design}
\label{sec:app_experimental_design}

All experiments only require CPUs (and no GPUs) and were computed on a Linux cluster (see Table~\ref{tab:cluster}). 

\begin{table}[ht]
\caption{Description of the infrastructure used for the experiments in this paper. }
\label{tab:cluster}
\centering
    \begin{tabular}{ll}
        \toprule
         \multicolumn{2}{c}{Computing Infrastructure} \\ \midrule
         Type & Linux CPU Cluster \\ 
         Architecture & 28-way Haswell-EP nodes \\
         Cores per Node & 1 \\
         Memory limit (per core) & 2.2 GB \\ \bottomrule
    \end{tabular}
\end{table}

The computational complexity of the PDP estimation with uncertainty is $\order\left(G \cdot n\right) \cdot \order(\hat c)$, with $\order(\hat c)$ being the runtime complexity of single surrogate prediction, $n$ denoting the size of the dataset to compute the Monte Carlo estimate and $G$ being the number of grid points. In the context of HPO, the general assumption is that the evaluation time of $\hat c$ is negligibly low as compared to evaluation $c$. So we argue that the runtime complexity of computing a PDP with uncertainty estimate can be neglected in this context. When computing ICE curves and their variance estimates beforehand, the algorithmic complexity of Algorithm \ref{alg:tree} corresponds to the algorithmic complexity of the tree splitting \citep{breiman:1984}. 

In our experiments, the runtimes to compute the PDPs and perform the tree splitting lies within a few minutes. We consider them to be negligible and will thus not report these. 

\subsubsection{BO on a Synthetic Function}
\label{sec:app_experimental_design_synth}

The Styblinski-Tang function 

\begin{eqnarray}
   c: \left[-5, 5\right]^d &\to& \R \\
   \lambdab &\mapsto& \frac{1}{2} \sum_{i = 1}^d \left(\lambdab_i^4 + 16 \lambdab_i^2 + 5 \lambdab_i\right)
\end{eqnarray}

was optimized via BO for $d \in \{3, 5, 8\}$ with a total budget of $\{80, 150, 250\}$ objective function evaluations, respectively, to allow sufficient optimization progress depending on the problem dimension. 

We computed an initial random design of size $4d$\footnote{The initial design was fixed across replications}. We performed BO with a GP surrogate model with a Mat\'{e}rn-3/2 kernel and the LCB acquisition function $a(\lambdab) = \hat m(\lambdab) + \tau \cdot \hat s(\lambdab)$ with different values $\tau \in \{0.1, 1, 5\}$. A nugget $10^{-8}$ was added for for numerical stability. We denote the best evaluated configuration, measured by $\hat c$, by $\hat \lambdab$. 

Based on the last surrogate model, we performed the partitioning in Algorithm \ref{alg:tree} for a total number of $5$ splits, with the different splitting criteria (see Section \ref{sec:app_experimental_results_mlp}), with PDPs being computed with a $G = 20$ equidistant grid points, and $n = 1000$ samples for the Monte Carlo approximation\footnote{Both grid-points and the data to compute the MC estimate are fixed across replications}. 

For all subsequent analysis, we considered the subregions $\Lambda^\prime$, for which $\hat \lambdab \in \Lambda^\prime$, and computed the PDPs according to Estimate \eqref{eq:pdp_gp_var_nocov}.

For every subregions considered $\Lambda^\prime$, we compute a partial dependence of $c$ for feature $\lambdab_1$, denoted as $c_1\left(\lambdab_1\right)$ to establish a ground-truth PDP estimate. 

\subsubsection{MLP}
\label{sec:app_experimental_design_mlp}

\begin{table}[]
\centering
    \caption{Hyperparameter space 1 of Auto-PyTorch Tabular.}
    \label{tab:searchspace}
    \begin{tabular}{cccc}
        \toprule
         Name & Range & log & type \\ \midrule
         Number of layers & $[1, 5]$ & no & int \\ 
         Max. number of units & $[64, 512]$ & yes & int \\
         Batch size & $[16, 512]$ & yes & int \\
         Learning rate (SGD) & $[1\textrm{e}^{-4}, 1\textrm{e}^{-1}]$ & yes & float \\
         Weight decay & $[1\textrm{e}^{-5}, 1\textrm{e}^{-1}]$ & no & float \\
         Momentum & $[0.1, 0.99]$ & no & float \\
         Max. dropout rate & $[0.0, 1.0]$ & no & float \\ 
         \bottomrule
    \end{tabular}
\end{table}

All experimental data were downloaded from the LCBench project\footnote{\url{https://github.com/automl/LCBench}, Apache License 2.0}. As an empirical performance model, we fitted a random forest (ranger) to approximate the relationship between hyperparameters and balanced error rate (BER). For every dataset, we performed a random search with $500$ iterations and evaluation via 3-fold cross-validation to choose reasonable for the hyperparameters represented in Table \ref{tab:random_forest_surr}. The empirical performance model acts as ground-truth in our experiments, and thus, we denote it by $c$. This function was used to compute the true PDP $c_S$.

\begin{table}[]
\centering
 \caption{Hyperparameter space of the random forest that was tuned over to compute the empirical performance model.}
\label{tab:random_forest_surr}
\begin{footnotesize}
    \begin{tabular}{cccc}
         \toprule
         Name & Range & log & type \\ \midrule
         Number of trees & $[10, 500]$ & yes & int \\
         mtry & $\{\textrm{true}, \textrm{false}\}$ & no & bool \\
         Minimum Size of Nodes & $[1, 5]$ & no & int \\
         Number of Random Splits & $[1, 100]$ & no & int \\ \bottomrule
    \end{tabular}
\end{footnotesize}
\end{table}

We computed an initial random design of size $2 \cdot d$\footnote{The initial design was fixed across replications}. We performed BO with a GP surrogate model with a Mat\'{e}rn-3/2 kernel and the LCB acquisition function $a(\lambdab) = \hat m(\lambdab) + \tau \cdot \hat s(\lambdab)$ with $\tau =1$. A nugget effect was modeled. The maximum budget per BO run was set to 200 objective function evaluations. We denote the best evaluated configuration, measured by $\hat c$, by $\hat \lambdab$. 

Based on the last surrogate model, we performed the partitioning in Algorithm \ref{alg:tree} for a total number of $6$ splits, with the different splitting criteria (see Section \ref{sec:app_experimental_results_mlp}), with PDPs being computed with a $G = 20$ equidistant grid points, and $n = 1000$ samples for the Monte Carlo approximation\footnote{The grid and the data used to compute the Monte Carlo estimate was fixed across replications}.

\subsection{Detailed Results }
\label{sec:app_experimental_results}

\subsubsection{Synthetic}
\label{sec:app_experimental_results_synth}

In Section \ref{sec:hpo_synth} we analyzed our tree-based partitioning method on the Styblinski-Tang function for different dimensions and degrees of sampling bias. To make the results of Table \ref{tab:tree_depth} more tangible, we visualized the associated PDPs and confidence bands for $\lambda_1$ and $\tau = 1$ of one iteration in Figure \ref{fig:pdp_synth}. The plots show clearly, that the number of splits required to obtain more confident and reliable PDP estimate in the sub-region containing the optimal configuration depends on the problem dimension. 

\begin{figure*}[ht]
    \centering
    \includegraphics[width=0.8\textwidth]{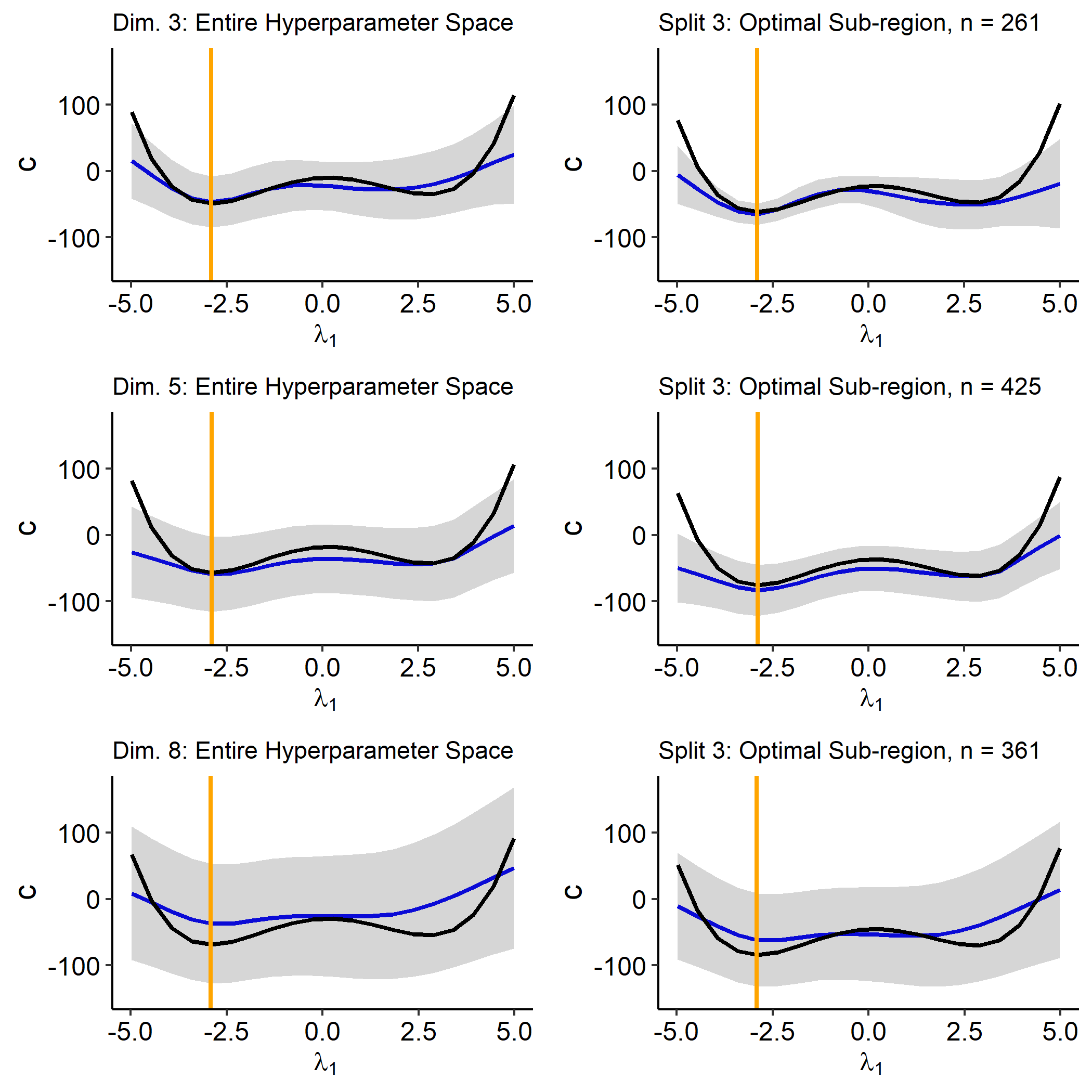}
    \caption{PDP (blue) and confidence band (grey) of the GP for hyperparameter $\lambda_1$ for the Styblinski-Tang function in case of 3 (top), 5 (middle) and 8 (bottom) dimensions. The black line shows the true PDP estimate of the Styblinski-Tang function. The orange vertical line marks the optimal configuration $\hat{\lambdab}_1$.}
    \label{fig:pdp_synth}
\end{figure*}

\subsubsection{MLP}
\label{sec:app_experimental_results_mlp}

In Section \ref{sec:hpo_nn} we evaluated the reliability of PDP estimation for the partitioning procedure proposed in Section \ref{sec:splitting}. The results presented in Section \ref{sec:hpo_nn} are aggregated over a total number of 35 different datasets. In Tables \ref{tab:conf.impr.dataset} and \ref{tab:loglik.impr.dataset} the relative improvement of the mean confidence (MC) and NLL are presented on dataset level. The mean and standard deviation are averaged over all hyperparameters. Furthermore, the mean values of the features providing the highest and lowest relative improvement for each dataset are reported. Following on that, Table \ref{tab:feat_best_worst} shows for each hyperparameter the number of datasets for which the respective hyperparameter led to the highest (lowest) relative improvement for both evaluation metrics. 
\begin{table}[t]
\centering
 \begin{scriptsize}
 \mbox{}\hfill
    \begin{minipage}[t]{.45\textwidth}
    \caption{Relative improvement of MC on dataset level. The table shows the mean ($\mu$) and standard deviation ($\sigma$) of the relative improvement (in $\%$) over all 7 hyperparameters and 30 runs after 6 splits. Additionally the mean value of the hyperparameter with the highest ($\mu_h$) and lowest ($\mu_l$) mean improvement are shown. }
    \label{tab:conf.impr.dataset}
         \begin{tabular}{lrrrr}
  \hline
  
Dataset & $\mu$ & $\sigma$ & $\mu_h$ & $\mu_l$ \\ 
   \hline
adult & 34 & 6 & 38 & 25 \\ 
  airlines & 49 & 20 & 61 & 3 \\ 
  albert & 57 & 26 & 78 & 14 \\ 
  Amazon\_employee\_access & 58 & 17 & 69 & 21 \\ 
  APSFailure & 46 & 17 & 60 & 22 \\ 
  Australian & 41 & 7 & 46 & 32 \\ 
  bank-marketing & 29 & 13 & 45 & 15 \\ 
  blood-transfusion-service & 34 & 20 & 39 & 13 \\ 
  car & 44 & 17 & 51 & 32 \\ 
  christine & 47 & 14 & 54 & 19 \\ 
  cnae-9 & 66 & 26 & 83 & 7 \\ 
  connect-4 & 47 & 14 & 56 & 17 \\ 
  covertype & 41 & 17 & 53 & 12 \\ 
  credit-g & 57 & 21 & 69 & 7 \\ 
  dionis & 49 & 21 & 63 & 5 \\ 
  fabert & 64 & 21 & 75 & 18 \\ 
  Fashion-MNIST & 41 & 12 & 47 & 18 \\ 
  helena & 43 & 16 & 52 & 8 \\ 
  higgs & 42 & 14 & 52 & 17 \\ 
  jannis & 35 & 13 & 44 & 19 \\ 
  jasmine & 46 & 11 & 56 & 27 \\ 
  jungle\_chess\_2pcs\_raw & 33 & 15 & 44 & 6 \\ 
  kc1 & 33 & 12 & 41 & 17 \\ 
  KDDCup09\_appetency & 52 & 21 & 63 & 3 \\ 
  kr-vs-kp & 46 & 14 & 56 & 26 \\ 
  mfeat-factors & 56 & 16 & 70 & 29 \\ 
  MiniBooNE & 36 & 14 & 42 & 18 \\ 
  nomao & 30 & 6 & 34 & 22 \\ 
  numerai28.6 & 60 & 28 & 76 & -3 \\ 
  phoneme & 29 & 7 & 32 & 25 \\ 
  segment & 53 & 21 & 66 & 10 \\ 
  shuttle & 48 & 11 & 58 & 32 \\ 
  sylvine & 37 & 6 & 42 & 29 \\ 
  vehicle & 34 & 8 & 41 & 30 \\ 
  volkert & 44 & 16 & 55 & 12 \\ 
   \hline
\end{tabular}
    \end{minipage} \hfill
    \begin{minipage}[t]{.45\textwidth}
     \caption{Relative improvement of the NLL on dataset level. The table shows the mean ($\mu$) and standard deviation ($\sigma$) of the relative improvement (in $\%$) over all 7 hyperparameters and 30 runs after 6 splits. Additionally the mean value of the feature with the highest ($\mu_h$) and lowest ($\mu_l$) mean improvement are shown. }
    \label{tab:loglik.impr.dataset}
      
        \begin{tabular}{lrrrr}
  \hline
Dataset & $\mu$ & $\sigma$ & $\mu_h$ & $\mu_l$ \\  
   \hline
adult & 13 & 6 & 23 & 8 \\ 
  airlines & 17 & 9 & 23 & 1 \\ 
  albert & 31 & 13 & 40 & 6 \\ 
  Amazon\_employee\_access & -0 & 36 & 29 & -35 \\ 
  APSFailure & 15 & 7 & 23 & 6 \\ 
  Australian & 12 & 14 & 23 & -4 \\ 
  bank-marketing & 7 & 9 & 17 & -1 \\ 
  blood-transfusion-service & 6 & 17 & 10 & -8 \\ 
  car & 26 & 32 & 35 & 10 \\ 
  christine & 10 & 11 & 17 & 1 \\ 
  cnae-9 & 67 & 37 & 93 & -11 \\ 
  connect-4 & -4 & 38 & 22 & -84 \\ 
  covertype & 28 & 13 & 38 & 8 \\ 
  credit-g & 41 & 24 & 81 & 2 \\ 
  dionis & 47 & 55 & 144 & -18 \\ 
  fabert & 37 & 17 & 54 & 8 \\ 
  Fashion-MNIST & 15 & 11 & 28 & 2 \\ 
  helena & -20 & 31 & -9 & -35 \\ 
  higgs & 20 & 12 & 33 & -2 \\ 
  jannis & 17 & 7 & 21 & 8 \\ 
  jasmine & 6 & 14 & 24 & -11 \\ 
  jungle\_chess\_2pcs\_raw & 9 & 15 & 24 & -7 \\ 
  kc1 & 11 & 10 & 17 & 4 \\ 
  KDDCup09\_appetency & 23 & 28 & 62 & -33 \\ 
  kr-vs-kp & 9 & 35 & 43 & -17 \\ 
  mfeat-factors & 25 & 19 & 51 & 10 \\ 
  MiniBooNE & 9 & 14 & 17 & -8 \\ 
  nomao & 8 & 6 & 16 & 3 \\ 
  numerai28.6 & 17 & 9 & 23 & 4 \\ 
  phoneme & 11 & 7 & 17 & 5 \\ 
  segment & 22 & 57 & 41 & -31 \\ 
  shuttle & 35 & 24 & 84 & 19 \\ 
  sylvine & 14 & 17 & 38 & -0 \\ 
  vehicle & 0 & 20 & 9 & -14 \\ 
  volkert & 23 & 18 & 40 & 5 \\ 
   \hline
\end{tabular}
    \end{minipage} \hfill
    \end{scriptsize}
\end{table}

\begin{table}[ht]
\centering
\caption{Number of datasets each of the hyperparameters had the highest $\mu_h$ and lowest $\mu_l$ average relative improvement w.r.t. MC and NLL. }
\label{tab:feat_best_worst}

\begin{footnotesize}

\begin{tabular}{ccccc}
\toprule 
&  \multicolumn{2}{c}{MC} &  \multicolumn{2}{c}{NLL}  \\ \cmidrule{2-3} \cmidrule{4-5} 
  
Hyperparameter & \# $\mu_h$ & \# $\mu_l$ & \# $\mu_h$ & \# $\mu_l$ \\ 
   \hline
Batch size &   1 &   3 &   3 &   4 \\ 
  Learning rate &   6 &   2 &   6 &   3 \\ 
  Max. dropout &   9 &   1 &   2 &   1 \\ 
  Max. units &   4 &  &   7 &  \\ 
  Momentum &   8 &  &   7 &   3 \\ 
  Number of layers &   3 &  14 &   9 &  11 \\ 
  Weight decay &   4 &  15 &   1 &  13 \\ 
   \bottomrule
\end{tabular}
\end{footnotesize}
\end{table}

\paragraph{Split Criteria}
In Section \ref{sec:splitting}, we introduced Eq.~\ref{eq:splitting_crit_L2} as split criteria within the tree-based partitioning of Algorithm \ref{alg:tree}. This measure is based on splitting ICE curves based on curve similarities, which is especially suitable in the underlying context as explained in Section \ref{sec:splitting}. 
However, we also compared it to two other measures that are based on uncertainty estimates provided by the probabilistic surrogate model.
The first one is also based on ICE curves of the variance function $\hat{s}^2(\lambdab_S, \lambdab_C^{(i)})$ and its PD estimate $\hat s^2_{S|\mathcal{N}^\prime}\left(\lambdab_S\right)$ within a sub-region $\mathcal{N}^\prime$. However, instead of minimizing the distance between curves and group the associated ICE curves regarding similar behavior, we can also minimize the area under ICE curves of the variance function. The reasoning for this is as follows: If we aim for tight confidence bands over the entire range of $\Lambda_S$, we want the ICE curves of the variance function to be - on average - as low as possible. This is equivalent to minimizing the average area under ICE curves of the variance function.
Thus, the calculation of Eq. \ref{eq:splitting_crit_L2} changes such that we first calculate the average area between each ICE curve of the uncertainty function and the respective sub-regional PDP
\begin{eqnarray*}
    L\left(\lambdab_S, i\right) = \frac{1}{G}  \sum\nolimits_{g = 1}^G\left(\hat s^2\left(\lambdab_S^{(g)}, \lambdab_C^{(i)(g)}\right) - \hat s^2_{S|\mathcal{N}^\prime} \left(\lambdab_S^{(g)}\right)\right),
\end{eqnarray*}
where $\hat s^2_{S|\mathcal{N}^\prime} \left(\lambdab_S^{(g)}\right):= \frac{1}{|\mathcal{N}^\prime|}\sum_{i \in \mathcal{N}^\prime} \hat s^2\left(\lambdab_S^{(g)}, \lambdab_C^{(i)(g)}\right)$ , and aggregate the quadratic value of it over all observations in the respective sub-region:
\begin{eqnarray}
    \mathcal{R}_{area}(\mathcal{N}^\prime) = \sum\nolimits_{i \in \mathcal{N}^\prime} L (\lambdab_S, i)^2. 
    \label{eq:splitting_crit_area}
\end{eqnarray}

Second, we also used the uncertainty estimates of the probabilistic surrogate model for each observation of the test data itself to define an impurity measure. Therefore we calculated the squared deviation of each observation to the mean uncertainty within the respective node. Hence, compared to the other two approaches, we do not group curves but the observations themselves regarding their uncertainty. We further refer to this approach as the \textit{variance (var)} approach.

As a third measure that is not based on the uncertainty estimates, we used the MSE of the posterior mean estimate of the surrogate model as split criterion. This is the most commonly used measure for regression trees and hence a solid baseline measure.

We compared the four impurity measures for the partitioning procedure over all datasets and hyperparameters. We compare the results that we presented in Section \ref{sec:hpo_nn} with the according results for the other three measures in Table \ref{tab:split_criteria_comp}. The impurity measure based on curve similarities that we used for our analysis (L2) outperforms the other three measures on average for all hyperparameters regarding MC and especially regarding OC. With regards to NLL there is not one measure that outperforms all others, but rather all measures perform on average over all hyperparameters equally good.

\begin{table*}[ht]
\caption{Comparison of different impurity measures regarding the relative improvement of MC, OC and NLL on hyperparameter level. The table compares the results of Table \ref{tab:conf.impr.feat} (L2) with the according results for the impurity measure based on Eq. \ref{eq:splitting_crit_area} (area), the \textit{variance} measure (var) and the \textit{mean} measure. }
\label{tab:split_criteria_comp}
\centering
\begin{footnotesize}
\begin{tabular}{ccccccccccccc}
 \toprule 
&  \multicolumn{4}{c}{$\delta$ MC (in \%)} &  \multicolumn{4}{c}{$\delta$ OC (in \%)} & \multicolumn{4}{c}{$\delta$ NLL (in \%)}  \\ \cmidrule{2-5} \cmidrule{6-9} \cmidrule{10-13}
Hyperparameter & L2 & area & var & mean & L2 & area & var & mean & L2 & area & var & mean \\ 
  \hline
Batch size & 41 & 40 & 38 & 36 & 62 & 58 & 55 & 53 & 20 & 19 & 16 & 19 \\ 
  Learning rate & 50 & 50 & 50 & 42 & 58 & 57 & 57 & 51 & 18 & 18 & 18 & 19 \\ 
  Max. dropout & 50 & 49 & 47 & 41 & 62 & 61 & 58 & 53 & 17 & 18 & 17 & 15 \\ 
  Max. units & 51 & 51 & 50 & 45 & 59 & 58 & 58 & 53 & 25 & 24 & 25 & 25 \\ 
  Momentum & 52 & 51 & 51 & 43 & 58 & 57 & 57 & 53 & 20 & 20 & 20 & 16 \\ 
  Number of layers & 31 & 30 & 29 & 25 & 51 & 46 & 46 & 45 & 14 & 15 & 15 & 13 \\ 
  Weight decay & 36 & 35 & 34 & 29 & 61 & 53 & 51 & 52 & 12 & 12 & 11 & 10 \\ 
   \bottomrule
\end{tabular}
\end{footnotesize}
\end{table*}

     

 \paragraph{Baseline comparison}
To emphasize the significance of our results we compare our results from Section \ref{sec:hpo_nn} with the following naive baseline method: We consider the L1-neighborhood around the optimal configuration -- where the GP can be assumed to be quite confident due to the focused sampling of BO -- which has the same size as the sub-region found by our method. We compute the PDP on this neighborhood, and compare it to the sub-regional PDP found by our method, in the same way as in Section \ref{sec:hpo_nn}. 
We calculated the average improvement of the three evaluation metrics over all datasets and repetitions on hyperparameter level as done in Table \ref{tab:conf.impr.feat} of our paper. While the mean confidence for our method improves on average by 30 - 52\%, the naive baseline method improves only by 8 - 23\%. Close to the optimal configuration, the mean improvement of our method is between 50-62\% while the baseline method only improves by 16-42\%. While the negative log-likelihood does not improve for the baseline method, it improves using our method by 12-24\%. See Table \ref{app:table_baseline_comp} for more detailed results. Hence, our method results in more reliable and confident PDP estimates than this baseline method.
These results justify using the more complex approach of grouping ICE curves based on similarity of their uncertainty structure to receive more reliable and confident PDP estimates in the sub-region close to the optimal configuration. Other disadvantages of the baseline method are that we need to specify the size of the neighborhood and that we only receive a rather local view around the optimal configuration. Our method on the other hand decomposes the global PDP in several distinct and interpretable sub-regions which helps the user to understand which regions of the entire hyperparameter space can be interpreted more reliably and which ones need to be regarded with caution. 

\begin{table*}[ht]
        \caption{Relative improvement of MC, OC and NLL on hyperparameter level. The table shows for our method and the baseline method the respective mean (standard deviation) of the average relative improvement over 30 replications for each dataset and 6 splits.}
        \label{app:table_baseline_comp}
        \centering
        \begin{footnotesize}
        \begin{tabular}{lrrrrrr}
         \toprule
         & \multicolumn{3}{c}{Tree-based partitioning} & \multicolumn{3}{c}{Baseline method}\\\cmidrule{2-7}
        Hyperparameter & $\delta$ MC (\%) & $\delta$ OC (\%) & $\delta$ NLL (\%) & $\delta$ MC (\%) & $\delta$ OC (\%) & $\delta$ NLL (\%) \\ 
         \midrule
        Batch size & 40.8 (14.9) & 61.9 (13.5) & 19.8 (19.5)& 13.7 (12.1) & 18.9 (16.0) & 1.4 (21.6)\\ 
        Learning rate & 50.2 (13.7) & 57.6 (14.4) & 17.9 (20.5)& 23.1 (17.7) & 27.2 (20.7) & -3.4 (27.0)\\ 
        Max. dropout & 49.7 (15.4) & 62.4 (11.9)  & 17.4 (18.2) & 21.1 (16.8) & 26.7 (16.8) & 3.3 (22.1)\\ 
        Max. units & 51.1 (15.2) & 58.6 (12.7) & 24.6 (22.0)& 19.1 (16.5) & 22.0 (17.1) & -1.4 (19.7)\\ 
        Momentum & 51.7 (14.5) & 58.3 (12.7)  & 19.7 (21.7)& 21.9 (16.4) & 25.3 (16.9) & 2.1 (25.4)\\ 
        Number of layers & 30.6 (16.4) & 50.9 (16.6) & 13.8 (32.5)& 8.1 (5.9) & 15.4 (12.8) & 0.9 (11.8)\\ 
        Weight decay & 36.3 (22.6) & 61.0 (13.1) & 11.9 (19.7)& 22.6 (15.9) & 41.7 (15.8) & 2.2 (24.2) \\ 
        \bottomrule
       
        \end{tabular}
        \end{footnotesize}
\end{table*}

 
       

\paragraph{Increased confidence with more splits} Furthermore, it needs to be noted that by using our method the mean confidence and NLL improve on average if we use six splits. However, this does not mean that they improve by design when splitting into sub-regions. As shown in Tables \ref{tab:conf.impr.dataset} and \ref{tab:loglik.impr.dataset}, improvements heavily depend on dataset and HP. Different factors influence the optimal number of splits, such as the sampling bias, size of the test-set, and dimensionality of the HP space.
For some of our benchmarks, the best results are reached with fewer splits. One example is shown in Figure \ref{fig:splitting_example}, where improvements in both metrics are made until Split 2 and by splitting deeper, estimates get less accurate especially when sample sizes in sub-regions become very small.
Thus, the number of splits is a (useful and flexible) control parameter in our method which can be determined within a human-in-the-loop approach (view plots after each split and stop when results are satisfying) or by defining a quantitative measure (e.g., based on a threshold for confidence improvement).  
\begin{figure}[!ht]
    \centering
    \includegraphics[width = 0.9\textwidth]{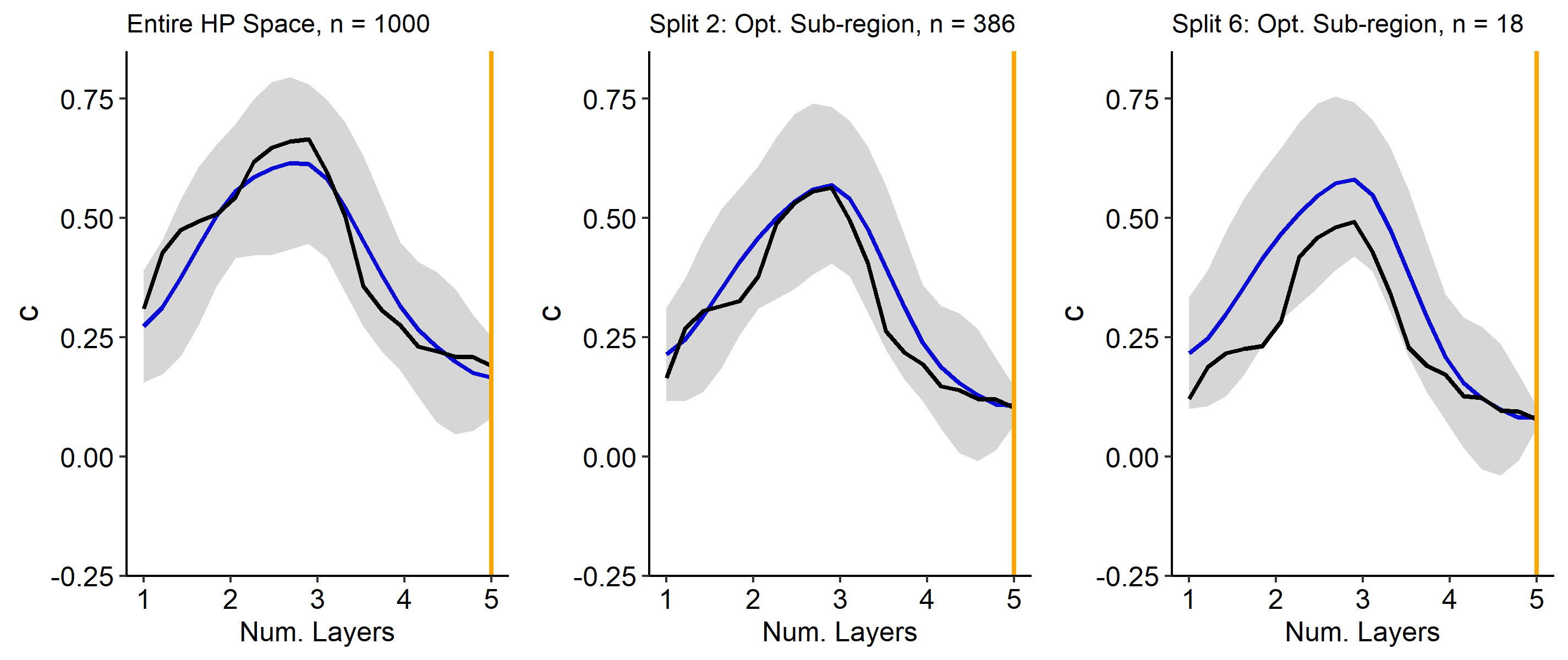}
    \caption{Estimated PDP of GP (blue) and true PDP estimate (black). The relative improvements after 2 (6) splits are $\delta$ \mbox{MC = $5\%$} ($0\%$) and $\delta$ NLL = $5\%$ ($-28\%$).}
    \label{fig:splitting_example}
\end{figure}
 
\section{Code}
\label{sec:code}

All code related to this paper is made available via a public repository\footnote{\url{https://github.com/slds-lmu/paper_2021_xautoml}}. All methods are implemented within the folder \texttt{R}, and all code used to perform the experiments are provided in \texttt{benchmarks}. All analyses shown in this paper in form of tables or figures can be reproduced via running the notebooks in \texttt{analysis}. 

\end{NoHyper}


\end{document}

%% file: latex-math/basic-math.tex
\ifdefined\N                                                                
\renewcommand{\N}{\mathds{N}}                                                
\else
  \newcommand{\N}{\mathds{N}}
\fi
\newcommand{\R}{\mathds{R}}                                                 
\ifdefined\C
  \renewcommand{\C}{\mathds{C}}                                             
\else
  \newcommand{\C}{\mathds{C}}
\fi

\def\argmin{\mathop{\sf arg\,min}}                                          
\newcommand{\order}{\mathcal{O}}                                            

\renewcommand{\P}{\mathds{P}}                                               
\newcommand{\E}{\mathds{E}}                                                 

%% file: latex-math/basic-ml.tex
\renewcommand{\xi}[1][i]{\mathbf{x}^{(#1)}}                                          

%% file: latex-math/ml-mbo.tex
\newcommand{\lambdab}{\boldsymbol{\lambda}}									